\pdfoutput=1

\documentclass[11pt]{article}
\makeatletter
\def\input@path{{latex/}{./}}
\makeatother
\usepackage[final]{acl}
\usepackage{booktabs}
\usepackage{color,soul}
\usepackage{algorithm}
\usepackage{algpseudocode}
\usepackage{svg}
\usepackage{geometry}
\usepackage{times}
\usepackage{tipa}
\usepackage{latexsym}
\usepackage{amsmath}
\usepackage{bm}
\usepackage{amsfonts}
\usepackage{cleveref}

\newcommand{\threestar}[1]{${\stackrel{***}{\text{\textbf{#1}}}}$}
\newcommand{\twostar}[1]{${\stackrel{**}{\text{\textbf{#1}}}}$}
\newcommand{\onestar}[1]{${\stackrel{*}{\text{\textbf{#1}}}}$}

\def\eqref#1{equation~\ref{#1}}

\def\1{\bm{1}}

\def\vh{{\bm{h}}}

\def\vv{{\bm{v}}}

\def\vy{{\bm{y}}}

\def\mH{{\bm{H}}}

\def\mY{{\bm{Y}}}

\DeclareMathAlphabet{\mathsfit}{\encodingdefault}{\sfdefault}{m}{sl}
\SetMathAlphabet{\mathsfit}{bold}{\encodingdefault}{\sfdefault}{bx}{n}

\def\sR{{\mathbb{R}}}

\usepackage{graphicx}
\usepackage{latexsym}
\usepackage{caption}
\usepackage[font=small,labelfont=bf]{subcaption}
\usepackage{booktabs}
\usepackage{enumitem}
\usepackage{multirow}
\usepackage{hyperref}
\usepackage{xcolor}
\usepackage{amsmath}
\usepackage{amsfonts}
\usepackage{amsthm}
\usepackage{scalefnt}
\usepackage{svg}
\usepackage{xspace}
\usepackage{fontawesome}
\usepackage{tcolorbox}
\usepackage{rotating}
\usepackage{subcaption}
\usepackage{dashrule}

\usepackage{pgfplots}
\usepackage{tikz}
\pgfplotsset{compat=1.18}
\usepackage{afterpage}

\usepackage{cleveref}
\crefname{section}{\S\!}{\S\S\!}
\crefname{table}{Table}{Tables}
\crefname{figure}{Figure}{Figures}
\crefname{algorithm}{Algorithm}{Algorithms}
\crefname{appendix}{Appendix}{Appendices}
\crefname{equation}{Equation}{Equations}
\crefname{lemma}{Lemma}{}
\Crefname{theorem}{Theorem}{}
\crefname{proposition}{Proposition}{}
\crefname{hypothesis}{Hypothesis}{}
\crefname{deduction}{Deduction}{}
\crefname{intuition}{\textbf{Intuition}}{\textbf{Intuitions}}
\crefname{observation}{\textbf{Observation}}{\textbf{Observations}}
\crefname{finding}{\textbf{Finding}}{\textbf{Findings}}
\crefname{cor}{Corollary}{Corollaries}

\usepackage{xspace}

\definecolor{tablerowcolor}{gray}{0.9}
\definecolor{mygray}{gray}{0.85}  

\definecolor{ForestGreen}{HTML}{009B55}
\definecolor{OrangeRed}{HTML}{ED135A}
\definecolor{CadetBlue}{HTML}{74729A}
\definecolor{SkyBlue}{HTML}{46C5DD}
\definecolor{SeaGreen}{HTML}{3FBC9D}
\definecolor{Peach}{HTML}{F7965A}
\definecolor{Goldenrod}{HTML}{FFDF42}
\definecolor{NavyBlue}{HTML}{006EB8}
\definecolor{Periwinkle}{HTML}{7977B8}
\definecolor{Orchid}{HTML}{AF72B0}
\definecolor{BlueViolet}{HTML}{473992}
\definecolor{SpringGreen}{HTML}{C6DC67}
\definecolor{ETHGray}{RGB}{0, 94, 184}	

\usepackage{collcell,fp}
\usepackage{colortbl} 
\usepackage{pgfmath} 

\usepackage{todonotes}
\usepackage{multirow}
\usepackage{array}
\usepackage{tcolorbox}
\definecolor{tticblue}{RGB}{0, 94, 184}  

\usepackage{xparse}

\NewDocumentCommand{\increase}{m m o}{%
    \IfValueTF{#3}{%
        ${#1}_{\textcolor{ForestGreen}{#2\uparrow}}^{\boldsymbol{#3}}$%
    }{%
        $#1_{\textcolor{ForestGreen}{#2\uparrow}}$
    }%
}%
\NewDocumentCommand{\decrease}{m m o}{%
    \IfValueTF{#3}{%
        ${#1}_{\textcolor{OrangeRed}{#2\downarrow}}^{\boldsymbol{#3}}$%
    }{%
        $#1_{\textcolor{OrangeRed}{#2\downarrow}}$%
    }%
}%

\newcommand{\interalia}[1]{\citep[\textit{inter alia}]{#1}}
\newcommand{\boldstart}[1]{\vspace{2pt}\noindent\textbf{#1}}

\usepackage[T1]{fontenc}

\usepackage{microtype}

\usepackage{inconsolata}

\usepackage{graphicx}
\usepackage{adjustbox}
\usepackage{booktabs}
\usepackage{multirow}
\usepackage[T1]{fontenc}
\usepackage[utf8x]{inputenc}
\title{
    How Tokenization Limits Phonological Knowledge Representation in Language Models and How to Improve Them
}

\author{
    Disen Liao$^{1,2}$ \quad Freda Shi$^{1,2}$ \\
    $^1$University of Waterloo \qquad $^2$ Vector Institute\\
    \texttt{\{d7liao,fhs\}@uwaterloo.ca}
}

\begin{document}
\maketitle

\begin{abstract}
    Tokenization is the first step in every language model (LM), yet it never takes the sounds of words into account. We investigate how tokenization influences text-only LMs' ability to represent phonological knowledge.
    Through a series of probing experiments, we show that subword-based tokenization systematically weakens the encoding of both local (e.g., rhyme) and global (e.g., syllabification) phonological features.
    To quantify this effect, we introduce the syllabification-tokenization alignment distance (STAD), a metric that measures the misalignment between a model's tokenization and the natural syllable boundaries of words, and find that higher misalignment correlates with poorer phonological representations, providing a simple diagnostic for phonology-aware tokenization.
    To address these limitations, we propose a lightweight IPA-based fine-tuning method that infuses phonological awareness into LMs, leading to consistent improvements across three phonology-related tasks while largely preserving math and general reasoning ability, with 1.1\% and 0.9\% drops on GSM8K and MMLU, respectively.\footnote{Our code is available at \url{https://github.com/liaodisen/Tokenization-Phonology}}
\end{abstract}


\section{Introduction}
Language models (LMs) have achieved remarkable success across diverse tasks, yet their understanding of sound structure, i.e., how words are pronounced, rhymed, or syllabified, remains poorly understood.
Surprisingly, even text-only LMs such as GPT-4o exhibit awareness of word pronunciation, enabling creative and educational applications like poetry generation \citep{zhang2024llm, yu2024charpoet} and language learning assistants \citep{hamaniuk2021potential, bonner2023large}.
This raises a fundamental question: how do LMs represent the phonological properties of words, despite never being trained on speech?
We focus on text-only models as they must infer phonology from orthography alone, which lets us isolate the role of text tokenization in that emergence.

\begin{figure}[!t]
    \centering
    \includegraphics[width=\linewidth]{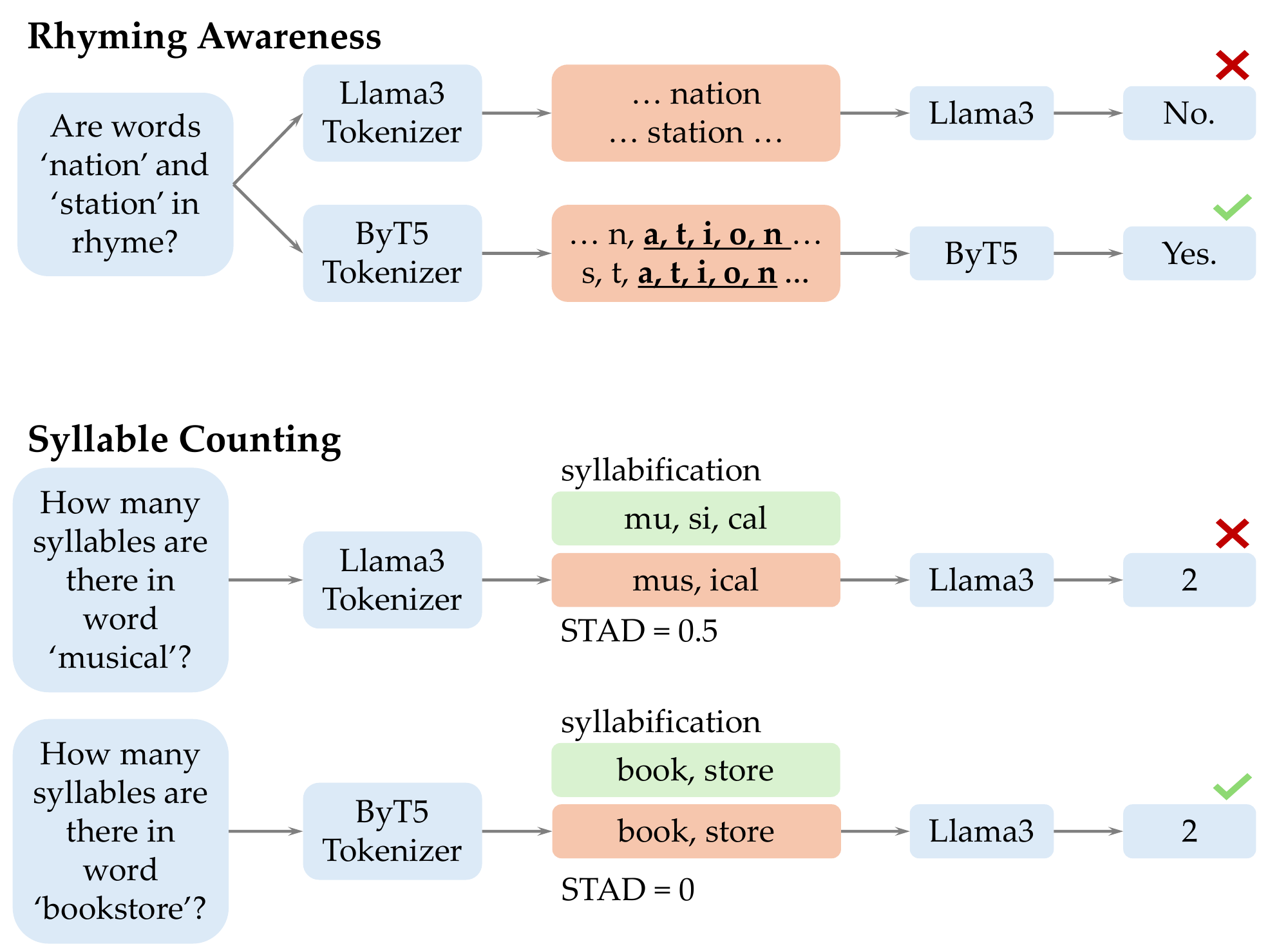}
    \caption{
        Illustrations of two key issues in LMs' phonological understanding caused by tokenization, using rhyming awareness and syllable counting as examples.
        The \textbf{top} shows that subword tokenization (e.g., BPE) may be too coarse to capture local phonological information, whereas the \textbf{bottom} shows that tokenization may misalign with natural syllable boundaries, leading to difficulties in learning prosodic structures.
    }
    \label{fig:phonology-explain}
    \vspace{-4pt}
\end{figure}

We take a systematic approach to examine how tokenization, the first stage of text processing in every LM, shapes the model's capacity to represent phonological structure.
We focus on two complementary aspects of phonological knowledge.
\begin{itemize}[leftmargin=*,itemsep=-2pt,topsep=0pt]
    \item \textbf{Local phonological features}, which capture whether words share similar sounds in part, for example, whether \textit{nation} and \textit{station} rhyme.
    \item \textbf{Prosodic structure}, which concerns the overall rhythmic organization of words, such as how \textit{musical} has three syllables and how graphemes map to phonemes in pronunciation.
\end{itemize}
To probe these aspects, we first consider three representative tasks---rhyming awareness, grapheme-to-phoneme (G2P) conversion, and syllable counting---and use linear probes to quantify how well hidden representations at different model layers encode relevant phonological features.

Our findings reveal that tokenization plays a pivotal role in shaping phonological competence.
Subword tokenization methods such as Byte Pair Encoding \citep{gage1994new,sennrich2015neural} and SentencePiece \citep{kudo2018subword}, while computationally efficient, are often (1) either too coarse to capture local phonological features or (2) misaligned with phonological boundaries (\cref{fig:phonology-explain}).
We further show that these issues systematically negatively affect models' ability to capture both rhyming relations and prosodic structure.
We introduce the syllabification-tokenization alignment distance (STAD) to quantify the misalignment between a model's tokenization and the natural syllable boundaries of words, and show that higher misalignment leads to degraded phonological representations.
This turns a broad intuition about tokenization into a measurable diagnostic that can guide tokenizer evaluation and design, while also clarifying how phonological structure can emerge in LMs despite the absence of speech.

To address these limitations, we propose a lightweight fine-tuning method that explicitly incorporates phonological information, offering a practical retrofit for existing models whose tokenizers cannot be easily changed.
By augmenting general-purpose QA data with phonological annotations, our approach enables LMs to reason more effectively about pronunciation while largely preserving their general reasoning ability, albeit with small trade-offs on broad benchmarks.

\section{Related Work}
\textbf{Probing the linguistic competence of LMs.}
Probing \citep{ettinger2016probing} investigates the internal representations of language models by training lightweight classifiers on hidden states to predict specific attributes, such as truthfulness \citep{azaria2023internal}, spatial understanding \citep{gurnee2023language}, sound perception \citep{ngo2024language}, and sound symbolism \citep{alper2024kiki}.
Compared to performance-based evaluation methods like accuracy, probing reveals more nuanced latent knowledge \citep[i.e., competence; ][]{chomsky1965aspects} embedded within internal activations \citep{burns2022discovering}---even when a model produces incorrect predictions, it may still encode relevant information.

Following recent work that advocates for using simple linear models as probes \citep{ettinger2016probing, alain2018understanding, hewitt2019structural}, in this work, we employ simple linear models to study how well the representations learned by LMs capture phonological features.

\boldstart{Phonology in LMs.}
Benchmarks for assessing the phonological capabilities of LMs are still in an early stage.
Recently, \citet{suvarna_phonologybench_2024} introduced a benchmark specifically designed to evaluate LLMs' performance on phonological tasks.
They proposed three tasks: rhyming generation, G2P, and syllable counting, to evaluate the phonology ability of LLMs, which motivates our choice of tasks in this work.

On the other hand, using LMs for phonological inference has become a promising direction, where some phoneme-based models have been tailored for lower-level phonological tasks.
For instance, PhonemeBERT \citep{sundararaman2021phoneme} is fine-tuned on a dataset combining ASR transcripts and phonemes, while Mix-Phoneme BERT \citep{zhang2022mixed} is pre-trained with phonemes and sub-phonemes as additional features to enhance text-to-speech performance.
Furthermore, \citet{qharabagh_llm-powered_2024} demonstrate that LLMs could significantly improve G2P conversion, especially in low-resource languages, showing the potential of LLMs to advance phonological processing in linguistically underserved contexts.
In contrast to this line of work, our study focuses on understanding how tokenization affects the phonological knowledge encoded in LMs' hidden states, aiming at offering fundamental insights that can inform future model design, as well as training, fine-tuning and inference strategies.

\boldstart{Tokenization Pitfalls.}
Subword-based tokenization algorithms such as BPE \citep{gage1994new,sennrich2015neural} are widely used in training modern LMs.
Prior work has highlighted how tokenization can introduce artifacts that impact model performance, particularly in tasks involving phoneme and grapheme representations \interalia{shin-etal-2020-autoprompt}.
Additionally, tokenization consistency plays a crucial role in extractive QA tasks \citep{sun-etal-2023-tokenization}.
\citet{singh2024tokenization} further argued that for numeric reasoning tasks, LLMs perform better when numbers are tokenized from right to left.
To mitigate tokenization-induced issues, we draw insights from \citet{deng2023rephrase} and \citet{bunzeck_small_2024}, and propose a fine-tuning strategy (\cref{sec:expr3}) that incorporates phonological information to help LMs overcome tokenization pitfalls.

Meanwhile, character-level tokenization has been explored as an alternative to subword-based methods to eliminate tokenization biases.
For instance, BERT has been shown to exhibit sensitivity to misspellings due to its reliance on subword tokenization \citep{sun2020adv}.
To address the limitations of subword tokenization, character-level or sub-character level tokenization strategies have been explored, including pre-training variants such as CANINE \citep{clark2022canine} and ByT5 \citep{xue2022byt5}.
In line with these existing efforts, our work highlights tokenizer pitfalls in phonological tasks, extending these findings toward better phonological understanding in LMs.

\section{Experiment 1: Probing Phonological Knowledge in LM Hidden States}
\label{sec:expr1}

We begin by addressing the research question \textbf{RQ1}: \textit{beyond performance-based evaluation} \citep{suvarna_phonologybench_2024}, \textit{how well do LMs encode phonological information in their hidden states?}

To answer this question, we adopt the probing approach in \citet{ettinger2016probing}.
Formally, given a language model $\mathcal{M}$, we prompt it with a text sequence $s$, tokenized into $|s|$ subwords $\{w_1, \dots, w_{|s|}\}$.
For each token $w_i$, at layer $\ell$, the model produces a hidden state vector $\vh_{i\ell} = \mathcal{M}(s, i, \ell) \in \mathbb{R}^{d}$, where $d$ is the hidden dimension of the model.
We take the hidden state at the last token at layer $\ell$ as the layer-wise representation of the entire prompt: $\vh_{\ell} = \vh_{|s|\ell}$.
This follows a common starting point in decoder-only probing work: prior studies on truthfulness/factuality, sentiment, and lexical structure also extract the final-token representation and train a lightweight probe on top of it, sometimes later comparing it against alternative pooling schemes \citep{burns2022discovering,liu-etal-2024-universal,palma-etal-2025-llamas,kaplan-etal-2025-tokens}.

Given a dataset of $n$ examples $\{s_1, \dots, s_n\}$, which have corresponding labels $\vy \in \mathbb{R}^n$ that represent certain phonological properties, we construct a probing feature set
\begin{equation}
    \bm{X} = [\vh^{(1)}_{\ell}, \dots, \vh^{(n)}_{\ell}]^\top \in \mathbb{R}^{n \times d},
    \label{label:eq:feature}
\end{equation}
where each $\vh^{(\cdot)}_{\ell}$ denotes the representation of prompt $s_j$ at layer $\ell$.
We then train a lightweight classifier or regressor (i.e., a probe) to predict $\vy$ from $\bm{X}$.
If the probe achieves performance above chance, it suggests that the LM's hidden states encode information relevant to the phonological property of interest \citep{hewitt2019designing}.
By applying this process across layers, we can analyze how phonological information evolves throughout the model hierarchy.

\begin{table*}[!t]
    \setlength{\tabcolsep}{3pt}
    \footnotesize
    \centering
    \renewcommand{\arraystretch}{1.2}
    \begin{tabular}{rrrrrrrrrrrrrrrrrrrr}
        \toprule
        \multicolumn{6}{c}{\bf Rhyming Awareness}
             &          & \multicolumn{6}{c}{\bf G2P}
             &          & \multicolumn{6}{c}{\bf Syllable Counting}                                                                                                                                                    \\
        \multicolumn{6}{c}{Accuracy (\%) by Layer Depth}
             &          & \multicolumn{6}{c}{$R^2$ by Layer Depth}
             &          & \multicolumn{6}{c}{$R^2$ by Layer Depth}                                                                                                                                                     \\
        \cmidrule(lr){1-6} \cmidrule(lr){8-13} \cmidrule(lr){15-20}
        0\%  & 20\%     & 40\%                                      & 60\%     & 80\%     & 100\% &  & 0\%   & 20\%     & 40\%  & 60\%     & 80\%  & 100\% &  & 0\%   & 20\%     & 40\%     & 60\%     & 80\%  & 100\% \\
        \midrule
        \rowcolor{mygray}
        \multicolumn{20}{l}{\textbf{BERT} (110M)}                                                                                                                                                                      \\
        56.0 & 67.6     & 68.3                                      & 70.9     & \bf 71.0 & 70.5  &  & .001  & \bf .073 & .003  & .007     & .031  & .060  &  & .020  & \bf .265 & .023     & .022     & .110  & .171  \\
        \rowcolor{mygray}
        \multicolumn{20}{l}{\textbf{GPT-2} (124M)}                                                                                                                                                                     \\
        63.4 & 64.7     & 66.1                                      & \bf 66.2 & 66.0     & 61.6  &  & .125  & \bf .188 & .180  & .159     & .113  & .145  &  & .476  & .624     & \bf .629 & .621     & .575  & .540  \\
        \rowcolor{mygray}
        \multicolumn{20}{l}{\textbf{GPT-neo-2.7B} (2.7B)}                                                                                                                                                              \\
        68.2 & 68.6     & \bf 72.4                                  & 69.6     & 69.7     & 67.0  &  & .083  & \bf .147 & .100  & -.035    & -.224 & .128  &  & .505  & \bf .662 & .631     & .514     & .464  & .492  \\
        \rowcolor{mygray}
        \multicolumn{20}{l}{\textbf{Mistral-7B-Instruct-v3} (7B)}                                                                                                                                                      \\
        64.5 & 80.6     & \bf 80.8                                  & 78.8     & 77.4     & 74.7  &  & .004  & .202     & .245  & \bf .315 & .276  & -.092 &  & .019  & .470     & .578     & \bf .582 & .560  & .312  \\
        \rowcolor{mygray}
        \multicolumn{20}{l}{\textbf{Llama3-8B-Instruct} (8B)}                                                                                                                                                          \\
        71.4 & \bf 80.7 & 76.6                                      & 76.8     & 70.5     & 78.4  &  & .041  & \bf .263 & .234  & .259     & .245  & .152  &  & .144  & \bf .661 & .660     & .634     & .563  & .532  \\
        \rowcolor{mygray}
        \multicolumn{20}{l}{\textbf{Llama3.1-8B-Instruct} (8B)}                                                                                                                                                        \\
        72.5 & \bf 79.8 & 79.0                                      & 77.9     & 77.3     & 74.9  &  & .052  & .330     & .346  & \bf .394 & .377  & .189  &  & .177  & \bf .713 & .703     & .710     & .685  & .616  \\
        \midrule
        \rowcolor{mygray}
        \multicolumn{20}{l}{\textbf{Control Experiment: Random Embeddings}}                                                                                                                                            \\
        48.8 & 48.7     & 51.7                                      & 49.3     & 50.2     & 50.8  &  & -.070 & -.100    & -.043 & -.073    & -.066 & -.082 &  & -.082 & -.073    & .001     & -.115    & -.097 & -.022 \\
        \bottomrule
    \end{tabular}
    \caption{
        Probe performance (accuracy for classification, and $R^2$ for regression) of different LM layers on the probing tasks averaged over 10 runs, higher is better.
        We take Layer $\lfloor\text{depth} \times \text{percentage}\rfloor$ for each depth percentage (0\%, ..., 100\%), where $0\%$ is the input embedding layer and $100\%$ is the final output.
        For each task, we boldface the performance of the best layer.
    }
    \label{table:main-vs-baselines}
\end{table*}

\subsection{Task Formulation and Setups}
Following \citet{suvarna_phonologybench_2024}, we select three phonology-related tasks that collectively capture both local phonological coherence and higher-level prosodic structures.
Each task is formulated as either a classification or regression problem, using input features $\bm{X}$ as defined in \cref{label:eq:feature}.

\boldstart{Task 1 - Rhyming Awareness} (classification): Given a pair of words, determine whether they form a perfect rhyme; that is, whether the words share the same stressed vowel and all subsequent sounds, regardless of spelling.
For example, \emph{night} (\textipa{/naIt/}) and \emph{kite} (\textipa{/kaIt/}) form a perfect rhyme because the stressed vowel \textipa{/aI/} and the following consonant \textipa{/t/} coincide exactly, whereas \emph{cough} (\textipa{/kOf/}) and \emph{tough} (\textipa{/t2f/}) do not, despite their orthographic similarity.
We fit a logistic regression that takes hidden states as the input to predict binary labels $y_i \in \{0, 1\}$.

To prevent models from exploiting superficial graphemic similarity, we restrict the dataset to rhyming pairs that differ in their final three orthographic characters, i.e., we will not include pairs such as \textit{procrastination} and \textit{verification}, which trivially rhyme due to shared suffixes.
Specifically, we compile 200 positive examples of perfect rhymes satisfying this condition and 200 negative examples of non-rhyming pairs, yielding a balanced dataset for binary classification.

\boldstart{Task 2 - Grapheme-to-Phoneme (G2P)} (regression): Convert a written word into its phonemic representation in ARPAbet \citep[e.g., \textit{cat} $\to$ \textit{K AE T}]{shoup1980phonological}.
We use the CMU Pronouncing Dictionary as the reference,\footnote{\url{http://www.speech.cs.cmu.edu/cgi-bin/cmudict}} which provides phonemic transcriptions of English words in ARPAbet notations.
In total, there are 39 possible phonemes used to describe English word pronunciations.
Each phoneme is encoded as a unique index from 0 to 39, with index 0 reserved for padding.
All phoneme sequences are padded to a fixed length of 8, the maximum number of phonemes per word in our dataset, yielding the output space $\mathbf{y}_i \in \mathbb{R}^8$.
For each LM, we train a multi-label ridge regression model using the hidden states as input features and the phoneme indices as targets.\footnote{
    ~We acknowledge that this task naturally fits better with a sequence generation formulation involving discrete sequential targets.
    However, such approaches typically require a more complex probe that risks storing linguistic knowledge within the probe itself, thereby confounding the interpretation \citep{voita2020information}.
    In addition, the limited size of our dataset makes training a complex model prone to overfitting.
    Therefore, we employ a simpler linear regression probe to both minimize potential information leakage and avoid overfitting.
}
The dataset consists of 2,000 words randomly sampled from the overlap between the CMU Pronouncing Dictionary and Google-10000-English, the most frequent 10,000 English words provided by Google.\footnote{\url{https://github.com/first20hours/google-10000-english}}

\boldstart{Task 3 - Syllable Counting} (regression): Predict the number of syllables in a given word.
We use the same dataset as in the G2P task, where each word is annotated with its syllable count using the same CMU-based procedure described by \citet{suvarna_phonologybench_2024}.
In this task, we fit a ridge regression model to predict the syllable count $y_i \in \mathbb{Z}_+$ from the hidden states.

\subsection{Results and Discussion}
We evaluate LMs with different architectures, sizes, and tokenization strategies, including BERT \citep{devlin2019bert}, GPT-2 \citep{radford2019language}, GPT-neo-2.7B \citep{gpt-neo}, Llama3-8B, Llama3.1-8B \citep{grattafiori2024llama}, Mistral-7B-v3 \citep{jiang2023mistral7b}.
We adopt the cross-validation paradigm, repeating each experiment 10 times with different random seeds and reporting the average performance.
In each experiment, we use 80\% of the data for training and the remaining 20\% for evaluation.
More details on probe training and hyperparameters can be found in \cref{app:control-exp,app:probing-details}.
Following \citet{hewitt2019designing}, we train probes with the same amount of random embedding examples as a control experiment to ensure that the probe is providing meaningful results.
The random embedding features are of the same dimension as the LM hidden states, but are sampled from a standard normal distribution.
Significantly outperforming random embedding--based probes indicates that there are indeed relevant features in the LM's hidden states.

According to \cref{table:main-vs-baselines}, with very few exceptions in G2P (e.g., BERT at Layer 0\% and GPT-neo-2.7B at Layers 60\% and 80\%), we find clear phonological signal in the hidden states of all tested LMs, as evidenced by their superior performance compared to the random embedding baseline across all tasks.
While we observe a general trend of larger models exhibiting stronger phonological knowledge, it is not universally true nor is it consistent across all tasks (for example, while Mistral-7B-Instruct-v3 achieves the best G2P performance overall, it underperforms smaller models like GPT-2 in syllable counting).

We also observe that phonological information is most prominently encoded in the middle layers of the models, typically between 20\% and 60\% of the total depth, indicating that prompting-based evaluations \citep{suvarna_phonologybench_2024} may underestimate the phonological knowledge present in LMs \citep{huPromptingNotSubstitute2023}.
However, the overall performance remains far from perfect, suggesting that there is still considerable room for improvement.

\section{Experiment 2: Explaining the Effects of Tokenization on Phonological Tasks}
\label{sec:expr2}
Having confirmed that LMs encode some imperfect phonological knowledge in their hidden states (\cref{sec:expr1}), we now turn to answering \textbf{RQ2}: \textit{do  tokenization strategies affect LMs' ability to capture phonological information, and if so, how?}

\subsection{Local Phonological Coherence: Exploiting Fine-Grained Tokenization for Rhymes}
\label{subsec:expr-local}
We hypothesize that subword-based tokenization may negatively impact an LM's ability to capture local phonological features, such as rhymes, due to its coarse granularity.
Therefore, we test the hypothesis by comparing the performance of LMs with standard tokenization against a finer-grained tokenization approach that splits words into individual characters separated by delimiters (e.g., by inserting slashes, the tokenization of \texttt{boy} becomes \texttt{b, /, o, /, y} instead of \texttt{boy}).
In the latter case, the model is forced to process each character separately, potentially allowing it to better capture phonological patterns like rhymes.
We use this character-level splitting only as a controlled diagnostic of granularity, not as a claim that character-level tokenization is the final design choice; a syllable-aware tokenizer would be a more phonologically grounded long-term direction.

\begin{table}[!t]
    \setlength{\tabcolsep}{3pt}
    \renewcommand{\arraystretch}{1.1}
    \footnotesize
    \centering
    \begin{tabular}{cc|cccccc}
        \toprule
        \textbf{Model} & \textbf{Format}
                       & \multicolumn{6}{c}{\textbf{Accuracy by Layer Depth}}                                                                                                                   \\
        \cmidrule(lr){3-8}
        \textbf{Size}  &                                                      & 0\%              & 20\%             & 40\%             & 60\%             & 80\%             & 100\%            \\
        \midrule \midrule
        \multicolumn{8}{c}{Subword Level Tokenization}                                                                                                                                          \\
        \midrule
        \rowcolor{mygray}
        \multicolumn{8}{l}{\textbf{BERT}}                                                                                                                                                       \\
        110M           & orig.                                                & 56.0             & 67.6             & 68.3             & 70.9             & 71.0             & 70.5             \\
                       & slash                                                & \threestar{68.6} & \twostar{74.5}   & \twostar{73.4}   & \twostar{77.5}   & \twostar{79.5}   & \twostar{78.1}   \\
        \rowcolor{mygray}
        \multicolumn{8}{l}{\textbf{GPT-2}}                                                                                                                                                      \\
        124M           & orig.                                                & 63.4             & 64.7             & 66.1             & 66.2             & 66.0             & 61.6             \\
                       & slash                                                & \threestar{71.7} & \threestar{76.9} & \threestar{77.2} & \threestar{79.1} & \threestar{78.5} & \threestar{77.5} \\
        \rowcolor{mygray}
        \multicolumn{8}{l}{\textbf{GPT-neo-2.7B}}                                                                                                                                               \\
        2.7B           & orig.                                                & 68.2             & 68.6             & 72.4             & 69.6             & 69.7             & 67.0             \\
                       & slash                                                & \textbf{73.2}    & \threestar{82.5} & \threestar{83.9} & \threestar{82.5} & \threestar{81.6} & \threestar{82.4} \\
        \rowcolor{mygray}
        \multicolumn{8}{l}{\textbf{Mistral-7B-Instruct-v3}}                                                                                                                                     \\
        7B             & orig.                                                & \textbf{64.5}    & 80.6             & 80.8             & 78.8             & 77.4             & 74.7             \\
                       & slash                                                & 55.8             & \textbf{81.1}    & \onestar{82.7}   & \textbf{79.5}    & \onestar{79.0}   & \twostar{77.6}   \\
        \rowcolor{mygray}
        \multicolumn{8}{l}{\textbf{Llama3-8B-Instruct}}                                                                                                                                         \\
        8B             & orig.                                                & \textbf{71.4}    & 80.7             & 76.6             & 76.8             & 70.5             & \textbf{78.4}    \\
                       & slash                                                & 56.3             & \onestar{85.4}   & \twostar{81.9}   & \textbf{77.9}    & \textbf{71.9}    & 76.1             \\
        \rowcolor{mygray}
        \multicolumn{8}{l}{\textbf{Llama3.1-8B-Instruct}}                                                                                                                                       \\
        8B             & orig.                                                & \textbf{72.5}    & 79.8             & 79.0             & 77.9             & 77.3             & 74.9             \\
                       & slash                                                & 56.3             & \twostar{85.1}   & \twostar{84.0}   & \textbf{80.0}    & \textbf{78.9}    & \twostar{79.5}   \\
        \midrule \midrule
        \multicolumn{8}{c}{Sub-Character Level Tokenization}                                                                                                                                    \\
        \midrule
        \rowcolor{mygray}
        \multicolumn{8}{l}{\textbf{ByT5-base}}                                                                                                                                                  \\
        580M           & orig.                                                & 45.5             & 79.6             & 81.0             & 79.9             & 80.3             & 66.3             \\
        \rowcolor{mygray}
        \multicolumn{8}{l}{\textbf{ByT5-small}}                                                                                                                                                 \\
        300M           & orig.                                                & 45.5             & 75.5             & 80.1             & 77.6             & 72.7             & 71.8             \\
        \midrule \midrule
        \rowcolor{mygray}
        \multicolumn{8}{l}{\textbf{Control Experiment: Random Embeddings}}                                                                                                                      \\
        -              & -                                                    & 48.8             & 48.7             & 51.7             & 49.3             & 50.2             & 50.8             \\
        \bottomrule
    \end{tabular}
    \caption{
        Accuracy (\%) of rhyming awareness probing tasks across models and layers.
        For each model, we report results using both the original tokenization (orig.) and slash-delimited tokenization (slash).
        The best performance for each model at each layer is indicated in boldface.
        In addition, a paired t-test is conducted to assess the hypothesis that slash-delimited tokenization yields higher accuracy than the original tokenization, with significance levels indicated as * ($p<0.05$), ** ($p<0.01$), and *** ($p<0.001$).
        Results with other delimiters (e.g., dot and comma) are with similar trends (\cref{app:delimiter}) and are omitted here for brevity.
    }
    \label{table:table_rhyme}
\end{table}

\boldstart{Results and Discussion.}
We fit logistic regression probes on the task of rhyming awareness and report accuracy with the same setup as in \cref{sec:expr1}.
In addition to models assessed in \cref{table:main-vs-baselines}, we also include ByT5-base and ByT5-small \citep{xue2022byt5}, which use finer-grained byte-level (i.e., sub-character level) tokenization strategies.

We find that inserting slashes to create a finer-grained tokenization improves the performance of all tested LMs across almost all layers, with statistical significance ($p < 0.05$) in most cases (\cref{table:table_rhyme}).
Extending the findings by \citet{zheng2025broken} that non-canonical tokenizations can be handled by LMs without harm to performance across text understanding tasks, our results suggest that such tokenization strategies can even enhance the model's ability to capture phonological features.

\begin{table*}[t!]
    \setlength{\tabcolsep}{3.4pt}
    \renewcommand{\arraystretch}{0.95}
    \footnotesize
    \centering
    \begin{tabular}{ccrrrrrrrrrrrrrr}
        \toprule
        \multirow{2}{*}{\textbf{Model}}                                      & \multirow{2}{*}{\textbf{STAD}} & \hspace{-2pt} & \multicolumn{6}{c}{\textbf{G2P} ($R^2$ by Layer Depth)} & \hspace{-2pt}    & \multicolumn{6}{c}{\textbf{Syllable Counting} ($R^2$ by Layer Depth)}                                                                                                                                                                                                    \\
        \cmidrule(lr){4-9} \cmidrule(lr){11-16}
                                                                             & \hspace{-2pt}                  &               & 0\%                                                     & 20\%             & 40\%                                                                  & 60\%               & 80\%             & 100\%            & \hspace{-2pt} & 0\%                & 20\%               & 40\%               & 60\%             & 80\%             & 100\%            \\
        \cmidrule(lr){1-2} \cmidrule(lr){4-9} \cmidrule(lr){11-16}
        \cellcolor{mygray}                                                   & .000 (A)                       & \hspace{-2pt} & .004                                                    & .056             & .001                                                                  & .002               & \textbf{.030}    & .009             & \hspace{-2pt} & {\threestar{.999}} & \threestar{.952}   & .009               & .022             & .129             & .054             \\
        \multirow{-2}{*}{\cellcolor{mygray} BERT}                            & .290 (M)                       & \hspace{-2pt} & \bf .009                                                & \bf.085          & .001                                                                  & .002               & .024             & \bf .010         & \hspace{-2pt} & .404               & .626               & \bf .068           & \bf .074         & \bf .264         & \bf .143         \\
        \midrule
        \cellcolor{mygray}                                                   & .000 (A)                       & \hspace{-2pt} & \threestar{.198}                                        & \threestar{.229} & {\threestar{.232}}                                                    & \threestar{.217}   & \threestar{.185} & \threestar{.194} & \hspace{-2pt} & .027               & {\threestar{.980}} & \threestar{.980}   & \threestar{.969} & \threestar{.952} & \threestar{.929} \\
        \multirow{-2}{*}{\cellcolor{mygray} GPT-2}                           & .388 (M)                       & \hspace{-2pt} & .081                                                    & {.148}           & .146                                                                  & .124               & .080             & .119             & \hspace{-2pt} & \bf .589           & {.740}             & .732               & .728             & .714             & .684             \\
        \midrule
        \cellcolor{mygray}                                                   & .000 (A)                       & \hspace{-2pt} & \threestar{.179}                                        & {\twostar{.219}} & \threestar{.211}                                                      & \threestar{.148}   & \threestar{.078} & \threestar{.004} & \hspace{-2pt} & \threestar{.945}   & \threestar{.953}   & {\threestar{.967}} & \threestar{.942} & \threestar{.930} & \threestar{.914} \\
        \multirow{-2}{*}{\cellcolor{mygray} GPT-neo-2.7B}                    & .388 (M)                       & \hspace{-2pt} & .072                                                    & {.169}           & .111                                                                  & .005               & -.124            & -.219            & \hspace{-2pt} & .555               & {.787}             & .758               & .692             & .634             & .539             \\
        \midrule
        \cellcolor{mygray}                                                   & .000 (A)                       & \hspace{-2pt} & \textbf{.001}                                           & .212             & {\textbf{.301}}                                                       & .314               & \textbf{.297}    & .239             & \hspace{-2pt} & .028               & \threestar{.800}   & {\threestar{.913}} & \threestar{.911} & \threestar{.854} & \threestar{.816} \\
        \multirow{-2}{*}{\cellcolor{mygray} Mistral-7B-Instruct-v3}          & .348 (M)                       & \hspace{-2pt} & .000                                                    & \bf .214         & .283                                                                  & \bf {.317}         & .282             & \bf .261         & \hspace{-2pt} & \bf .045           & .708               & .804               & {.806}           & .789             & .762             \\
        \midrule
        \cellcolor{mygray}                                                   & .000 (A)                       & \hspace{-2pt} & \threestar{.034}                                        & \threestar{.349} & \threestar{.356}                                                      & {\threestar{.370}} & \threestar{.366} & \threestar{.325} & \hspace{-2pt} & .152               & \threestar{.931}   & {\threestar{.935}} & \threestar{.923} & \threestar{.899} & \threestar{.860} \\
        \multirow{-2}{*}{\cellcolor{mygray} Llama3-8B-Instruct}              & .372 (M)                       & \hspace{-2pt} & .023                                                    & .295             & .297                                                                  & {.333}             & .308             & .276             & \hspace{-2pt} & \bf.165            & .769               & {.795}             & .769             & .749             & .717             \\
        \midrule
        \cellcolor{mygray}                                                   & .000 (A)                       & \hspace{-2pt} & \textbf{.033}                                           & \onestar{.325}   & \twostar{.321}                                                        & {\twostar{.387}}   & \twostar{.357}   & \textbf{.166}    & \hspace{-2pt} & .188               & \threestar{.936}   & {\threestar{.939}} & \threestar{.921} & \threestar{.898} & \threestar{.859} \\
        \multirow{-2}{*}{\cellcolor{mygray} Llama3.1-8B-Instruct}            & .372 (M)                       & \hspace{-2pt} & .029                                                    & .304             & .285                                                                  & {.349}             & .317             & .157             & \hspace{-2pt} & \bf.211            & .783               & {.789}             & .769             & .754             & .723             \\
        \midrule

        \multicolumn{2}{l}{\cellcolor{mygray} Control: Randomized Embedding} & \hspace{-2pt}                  & -.070         & -.101                                                   & -.043            & -.073                                                                 & -.066              & -.082            & \hspace{-2pt}    & -.082         & -.073              & .001               & -.115              & -.097            & -.022                               \\
        \bottomrule
    \end{tabular}
    \caption{
        $R^2$ for G2P and syllable counting, comparing words with aligned (A; STAD = 0) vs. misaligned (M; STAD > 0.25) tokens and syllables.
        For each model and layer, the best $R^2$ is presented in boldface.
        A one-sided $t$-test is performed between the A and M splits at each layer to evaluate whether aligned words yield significantly better performance, with significance levels denoted as * ($p < 0.05$), ** ($p < 0.01$), and *** ($p < 0.001$). Results for more models can be found in the Appendix.
    }
    \label{table:stad}
\end{table*}

\subsection{Alignment between Tokens and Syllables Captures Prosodic Structure}
\label{subsec:expr-global}
Prosodic structure refers to the rhythmic and hierarchical organization of spoken language, including units such as syllables, stress patterns, and intonation.
These features jointly shape how words are pronounced beyond their written form.
To examine how tokenization may implicitly capture such prosodic regularities, we focus on two diagnostic tasks, G2P and syllable counting, as representative proxies for prosodic structure.

We hypothesize that closer alignment between tokenization and syllabification corresponds to better performance on phonological tasks.
To evaluate this hypothesis, we introduce a metric, the \textit{syllabification--tokenization alignment distance} (STAD), which quantifies the degree of mismatch between tokenization and syllabification.
We then compare the performance of language models (LMs) on words with high versus low STAD values.

\boldstart{STAD.}
Tokenizers segment words using frequency-based heuristics \citep{gage1994new,kudo2018sentencepiece}, which often diverge from syllable boundaries because phonological principles are not explicitly encoded in the process.
The STAD metric measures this divergence, with higher values indicating greater misalignment.
Consider a word with $n+1$ characters $w = a_1a_2 \dots a_{n+1}$, where there are $n$ possible positions to insert a split.
We represent tokenization and syllabification as two vectors, $\vv_{\text{tok}} = [b_1, b_2, \dots, b_n]$ and $\vv_{\text{syl}} = [c_1, c_2, \dots, c_n]$, where each $b_i, c_i \in \{0, 1\}$ indicates if a split occurs after the $i$-th character (1) or not (0).
The STAD is defined as the normalized Hamming distance between $\vv_{\text{tok}}$ and $\vv_{\text{syl}}$:
\vspace{-5pt}
\begin{align*}
    \text{STAD}(\vv_{\text{tok}}, \vv_{\text{syl}}; w) = \frac{\sum_{i=1}^n |b_i - c_i|}{n}.
\end{align*}
Taking the word \texttt{musical} as an example: it is syllabified as \texttt{[`mu', `si', `cal']}, represented by $\vv_{\text{syl}} = [0, 1, 0, 1, 0, 0]$.
Meanwhile, if a tokenizer splits it as \texttt{[`mus', `ical']}, yielding $\vv_{\text{tok}} = [0, 0, 1, 0, 0, 0]$, the STAD becomes
\begin{align*}
    \frac{0 + 1 + 1 + 1 + 0 + 0}{6} = 0.5.
\end{align*}
We obtain reference syllable boundaries using \texttt{syllabify}\footnote{\url{https://github.com/eoleedi/syllabify}}, a CMU Pronouncing Dictionary--based English syllabification toolkit consistent with the CMU-based preprocessing used in \citet{suvarna_phonologybench_2024}.
For each model, token boundaries are obtained by running that model's own tokenizer on the word and projecting the resulting splits back to character positions.

\boldstart{Data and experimental setup.}
We evaluate the same set of LMs with subword-level tokenization as in \cref{sec:expr1}.
For each LM, we create two splits of words, the token-syllable aligned (A; STAD = 0) split and the token-syllable misaligned (M; STAD $>$ 0.25) one, with samples of 1,000 words in each split.
We train ridge regression probes on the hidden states of each layer for both G2P and syllable counting tasks, following the same setup as in \cref{sec:expr1}.

\boldstart{Results and Discussion.}
For both tasks, all LMs except BERT and Mistral-7B-Instruct-v3 show significantly better performance on words with low STAD scores at layers deeper than 20\% of the model depth under the paired one-sided $t$-tests reported in \cref{table:stad}, suggesting that tokenization structure influences the models' ability to capture prosodic features.
Generally, better alignment between tokenization and syllabification (i.e., lower STAD) corresponds to better performance on phonological tasks.
Even for Mistral-7B-Instruct-v3, which shows no clear trend in the G2P task, performance on low-STAD words is consistently higher in the syllable counting task.

We hypothesize that the absence of a clear pattern in BERT arises from its distinct architecture---specifically, its use of bidirectional attention, which differs from the autoregressive setup of all the other models.
We leave a detailed examination of this phenomenon to future work.

\begin{figure}[!t]
    \centering
    \includegraphics[width=1\linewidth]{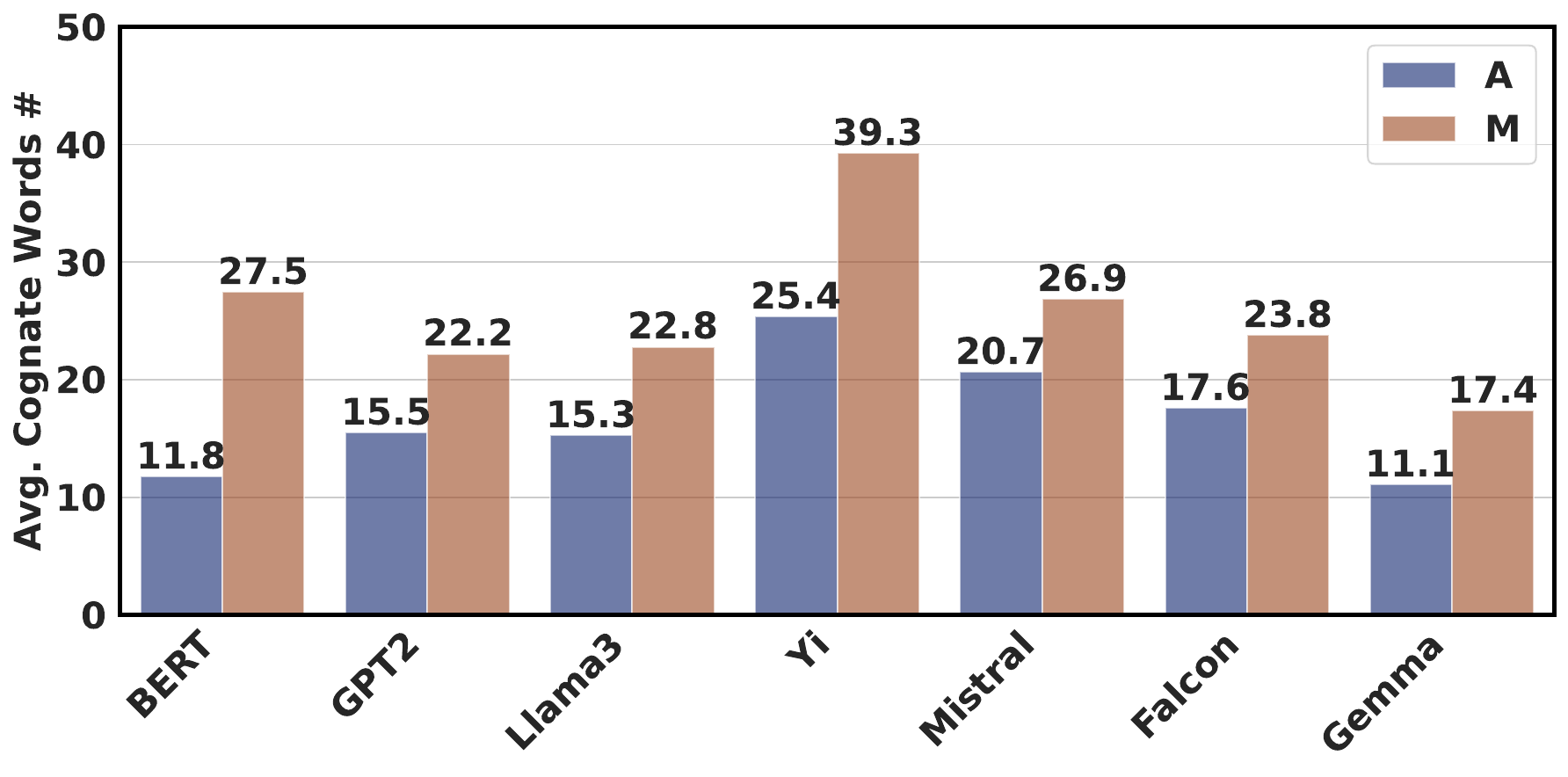}
    \caption{Average number of CogNet-related entries for token-syllable aligned words (A; STAD $=$ 0) and token-syllable misaligned words (M; STAD $>$ 0.25) for different tokenizers.}
    \label{fig:cognate}
\end{figure}

\subsection{Conjecture: Loanwords and Cognates Cause Syllabification--Tokenization Misalignment}
\label{subsec:expr-cognate}
We propose a possible explanation for why certain words exhibit tokenizations that misalign with their syllabifications and, consequently, are more difficult for LMs to process in phonological tasks.
These words may tend to display high orthographic variability in the training corpus, often stemming from historical factors such as lexical borrowing or etymological divergence.
When a word is shared across languages, it is often adapted to different spelling conventions (e.g., \textit{music} vs. \textit{musik}), increasing orthographic variability in the training corpus.
As a result, such words may contain letter $n$-grams that occur infrequently in the tokenizer's training data, leading subword-based tokenizers to produce segmentations that deviate from natural syllable boundaries.
Taking these together, we hypothesize that words with high cross-linguistic relatedness are more likely to be tokenized in ways that misalign with their syllabification.

We use CogNet \citep{batsuren2019cognet}, a comprehensive database of cognate words and loanwords, to estimate the level of cross-linguistic relatedness for each word in our dataset.
Specifically, we use the number of related entries listed in CogNet as a proxy for this relatedness, and present the average number of such entries for words in the aligned (A) and misaligned (M) groups (\cref{fig:cognate}).
In addition to tokenizers of aforementioned models, we also include three additional tokenizers, Yi \citep{young2024yi}, Falcon \citep{almazrouei2023falconseriesopenlanguage}, and Gemma \citep{team2024gemma}.
The results clearly indicate that words in group M systematically have a higher number of cognate or loanword variants than those in group A across all evaluated tokenizers.
This finding suggests that words with extensive cross-linguistic variation are more likely to undergo tokenization that misaligns with their syllabification, thereby complicating the model's ability to capture their phonological features.

In summary, our experiments in this section demonstrate that tokenization strategies substantially influence LMs' ability to capture both local phonological features (e.g., rhymes) and global prosodic structures (e.g., syllable structures).
These findings suggest that STAD can serve as a simple diagnostic for evaluating phonology-aware tokenization, and that tokenization strategies that better align with phonological principles may lead to improved phonological reasoning in LMs.

\section{Experiment 3: Fine-tuning LMs for Better Phonological Reasoning}
\label{sec:expr3}
\begin{figure*}[!t]
    \centering
    \begin{subfigure}{0.32\textwidth}
        \centering
        \includegraphics[height=130pt]{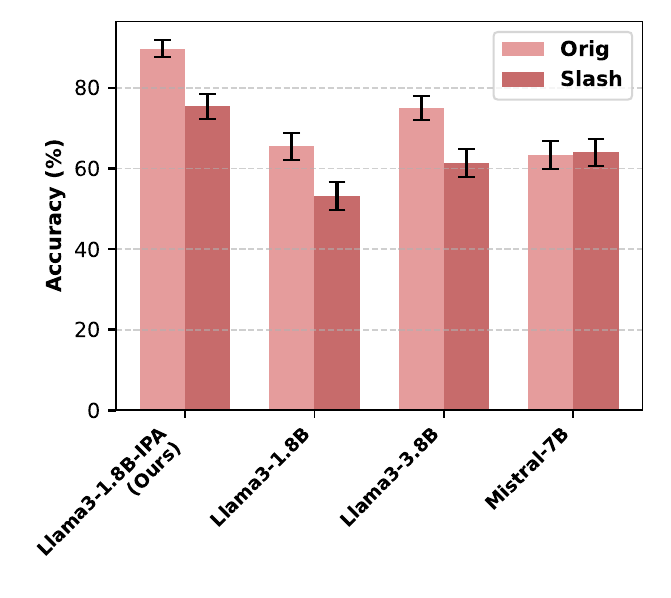}
        \vspace{-24pt}
        \caption{Rhyming Awareness.}
    \end{subfigure}
    \hfill
    \begin{subfigure}{0.32\textwidth}
        \centering
        \includegraphics[height=130pt]{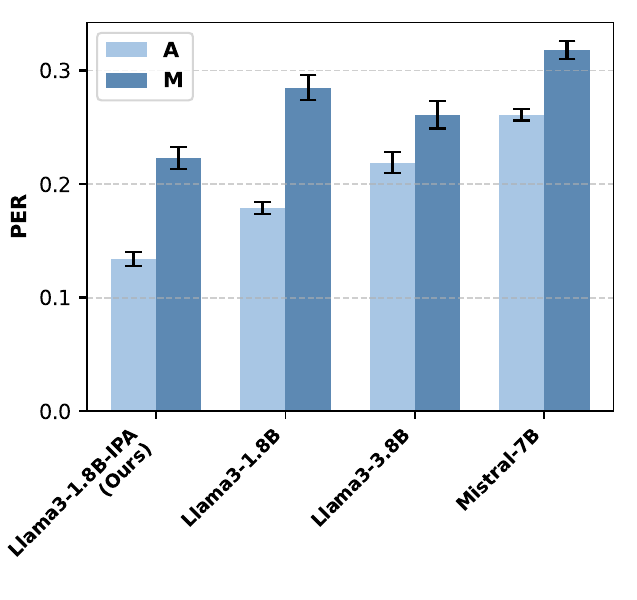}
        \vspace{-10pt}
        \caption{G2P.}
    \end{subfigure}
    \hfill
    \begin{subfigure}{0.32\textwidth}
        \centering
        \includegraphics[height=130pt]{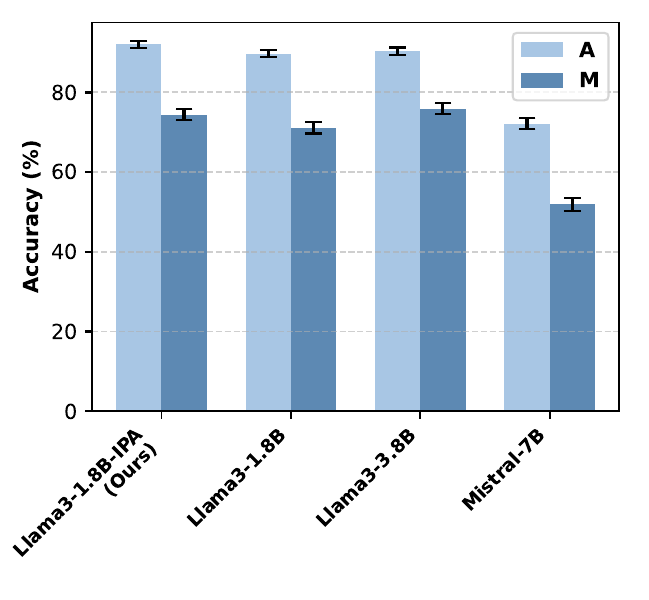}
        \vspace{-10pt}
        \caption{Syllable Counting.}
    \end{subfigure}
    \caption{Performance evaluated on three phonology-related tasks.}
    \label{fig:finetune}
\end{figure*}

Having revealed that tokenization materially affects how LMs internally encode phonological features through probing analyses, we now move toward a practical question.
Specifically, we seek to address \textbf{RQ3}: \textit{How can we enhance LMs' phonological reasoning while largely preserving their general language understanding?}

We propose a simple yet effective strategy: fine-tuning LMs with explicit phonological awareness through a carefully curated instruction-tuning dataset.
While designing a syllable-aware tokenizer would be a more direct architectural solution, replacing the tokenizer of an existing foundation model is substantially more invasive than post-training.
Our goal here is therefore to provide a practical patch for widely used text-only LMs without modifying their vocabulary.
The International Phonetic Alphabet \citep[IPA;][]{international1999handbook} provides a standardized and transparent representation of word pronunciation, offering direct cues to syllable structure and rhyme.
However, prior work shows that even though LMs possess partial knowledge of IPA, they struggle to leverage it effectively in phonology-related tasks \citep{suvarna_phonologybench_2024}.

To overcome this limitation, we introduce a data augmentation approach that integrates IPA transcriptions into general-purpose QA examples and fine-tunes LMs to reason explicitly over IPA in phonology-related tasks, while maintaining exposure to general tasks to mitigate catastrophic forgetting \citep{kirkpatrick2017overcoming}.
For general QA data, we randomly sample 0--2 words per question and enclose them with IPA indicator tokens (i.e., \texttt{<IPA>} and \texttt{</IPA>}, inserted as two tokens into the vocabulary) to signal the model to utilize IPA information.
The model's answer is then prefixed with the IPA transcription of the selected words; if no words are sampled, the example remains unchanged.
Even when IPA is not semantically necessary for the QA content itself, this interleaving serves a dual purpose: it preserves exposure to the original instruction-following distribution while teaching the model to attend to phonological side information in parallel with semantics.
For phonology-specific tasks, we construct conversational examples in which the answer explicitly reasons over IPA transcriptions (see \cref{app:data example,fig:conversation-example,fig:rhyme-train,fig:g2p-train,fig:syllable-train} for detailed templates and examples).

\boldstart{Fine-tuning setups.}
Our dataset for fine-tuning consists of two sources: (1) high-quality general instruction-tuning conversation data sampled from OpenHermes2.5 \citep{OpenHermes2.5}, and (2) phonology-related tasks, where we select words out of Google-10000-English as training examples to inform the model about using IPA for the phonology-related tasks, leaving the evaluation set completely unseen during training.
We fine-tune Llama3.1-8B-Instruct using low-rank adaptation \citep[LoRA;][]{hu2021loralowrankadaptationlarge} on our constructed instruction-tuning dataset (with statistics presented in \cref{tab:fine-tune data}), and denote it as Llama3.1-8B-IPA.
For evaluation, we assess three strong baselines from instruction-tuned LLMs, Llama3.1-8B-Instruct, Llama3-8B-Instruct, and Mistral-7B-Instruct-v3. Additional fine-tuning and evaluation details are provided in \cref{app:finetune-details}.

\begin{table}[!t]
    \setlength{\tabcolsep}{5pt}
    \small
    \centering
    \begin{tabular}{lr}
        \toprule
        \textbf{Data Type} & \textbf{Number of Examples} \\
        \midrule
        Conversation       & 3,000                       \\
        Rhyming Awareness  & 200                         \\
        Syllable Counting  & 500                         \\
        G2P                & 500                         \\
        \bottomrule
    \end{tabular}
    \caption{Number of fine-tuning examples from each source.}
    \label{tab:fine-tune data}
\end{table}

\begin{table}[t]
    \setlength{\tabcolsep}{5pt}
    \small
    \centering
    \begin{tabular}{lcc}
        \toprule
        \textbf{Model}         & \textbf{GSM8K} & \textbf{MMLU}  \\
        \midrule
        Llama3.1-8B-Instruct   & 69.9 $\pm$ 0.4 & 65.3 $\pm$ 0.5 \\
        Llama3.1-8B-IPA (Ours) & 68.8 $\pm$ 0.5 & 64.4 $\pm$ 0.6 \\
        \bottomrule
    \end{tabular}
    \caption{Performance of Llama3.1-8B-Instruct and the fine-tuned Llama3.1-8B-IPA model on general-purpose reasoning tasks.
        Metrics are accuracy (mean $\pm$ standard deviation) across 3 inference runs.}
    \label{tab:other task}
    \vspace{-4mm}
\end{table}

\boldstart{Evaluation metrics.}
We evaluate the same tasks using the same datasets described in \cref{sec:expr1}, conducting zero-shot inference, with prompt templates provided in the appendix.
However, unlike the probing experiments, we now conduct performance-based evaluations, by directly instructing the models to output answers.
We report the accuracies for the rhyming awareness and syllable counting tasks, and the phoneme error rate (PER; the Levenshtein distance between the predicted and reference over the number of phonemes in the reference) for the G2P task.
Lower PER indicates better performance.

To ensure there is no catastrophic forgetting in our fine-tuned model, we also evaluate it on two widely used reasoning benchmarks: GSM8K \citep{cobbe2021trainingverifierssolvemath} and MMLU \citep{hendrycks2021measuringmassivemultitasklanguage}.
We employ a chat-style zero-shot evaluation to simulate real-world user interactions.
For MMLU, we randomly sample three subjects from each major category (STEM, social sciences, humanities, and other), totaling 12 subjects.
Results are averaged over 3 runs using decoding parameters \texttt{top-p}=0.95, \texttt{temperature}=0.8.

\boldstart{Results and discussion.}
As presented in \cref{fig:finetune}, fine-tuning on the IPA-augmented dataset consistently improves performance across all three phonology-related tasks.
For the two prosodic structure tasks (G2P and syllable counting), LMs perform considerably better on words with low STAD scores (A), indicating that tokenization structure also affects inference performance, consistent with our findings in representation-level probing (\cref{sec:expr2}).
For the rhyming awareness task, however, we find that simply inserting slashes into words does not necessarily improve performance, suggesting that tokenization alone is not sufficient to enhance phonological reasoning at the performance level.

Despite 1.1\% and 0.9\% drops on GSM8K and MMLU, respectively, our fine-tuned model, Llama3.1-8B-IPA, retains most of its general reasoning and knowledge abilities (\cref{tab:other task}) without clear signs of catastrophic forgetting.
The trade-off is therefore not zero-cost, but favorable for our goal of specialized phonological adaptation.
Our objective here is not to improve general reasoning benchmarks per se, but to show that interleaving IPA signals with general instruction data can inject phonological knowledge while maintaining most general capability.
This may be attributed to the domain mismatch between the fine-tuning dataset and the evaluation benchmarks, our limited hyperparameter tuning due to limited compute budget, as well as the relatively small capacity of the base model (compared to the commercial ones).
We remain fairly optimistic that other advanced techniques, such as reinforcement learning--based ones, can be explored to further mitigate forgetting while enhancing phonological reasoning.

\section{Conclusion and Discussion}
We analyze a set of LMs on three tasks that cover both local and hierarchical phonological structures, and show that tokenization can introduce systematic biases in word representation and thereby limit phonological performance.
More broadly, these results suggest that phonological failures in LMs need not reflect a complete absence of phonological knowledge; much of the difficulty appears to arise at the interface between subword segmentation and phonological structure.
STAD turns this observation into a concrete diagnostic for tokenizer evaluation, and our cognate analysis suggests one linguistic source of such misalignment, though the full causal relationship remains an open question.
Together with the fine-tuning results, these findings motivate two complementary paths forward: tokenizer design that better respects phonological boundaries, and lightweight post-training with explicit phonological supervision.

These lessons may also inform joint speech and text language models \citep[e.g., ][\emph{inter alia}]{chou2023toward}, by highlighting the importance of text tokenizers that preserve phonological structure, though direct comparisons with speech or multimodal systems remain future work.
We do not see risks beyond the minimal risks of any research in computational linguistics.

\section*{Limitations}
First, it is worth noting that our experiments have been primarily focused on English, a representative alphabetic language.
Findings in this work need significant work to be possibly adaptable to logographic languages.
We leave the exploration of a broader range of modal architectures and additional languages for future work.

Second, most of the analytical experiments in this work are correlational in nature, and we acknowledge that there are many confounding factors that may affect the causality of the observed effects, particularly when it comes to hypotheses related to cognates.
We encourage future work to further investigate these relationships.

Finally, while our proposed fine-tuning method has demonstrated effectiveness in enhancing phonological understanding, it is important to note that the improvements are not uniform across all evaluated models.
In addition, due to the compute constraint, there is no guarantee that findings from this work generalize to larger, proprietary models.
We leave a more comprehensive evaluation of the proposed analysis (\cref{sec:expr1,sec:expr2}) and method (\cref{sec:expr3}) on a wider range of models to future work.

\section*{Acknowledgment}
We thank three anonymous reviewers and the area chair for the valuable suggestions.
This work is supported in part by an NSERC Discovery Grant (RGPIN-2024-04395) and a Canada CIFAR AI Chair Award to FS. 


\bibliography{custom}
\clearpage
\appendix

\section{Cognates, Loanwords \& CogNet}
\label{app:cognate_examples}

\begin{figure}[!h]
    \centering
    \includegraphics[width=0.8\linewidth]{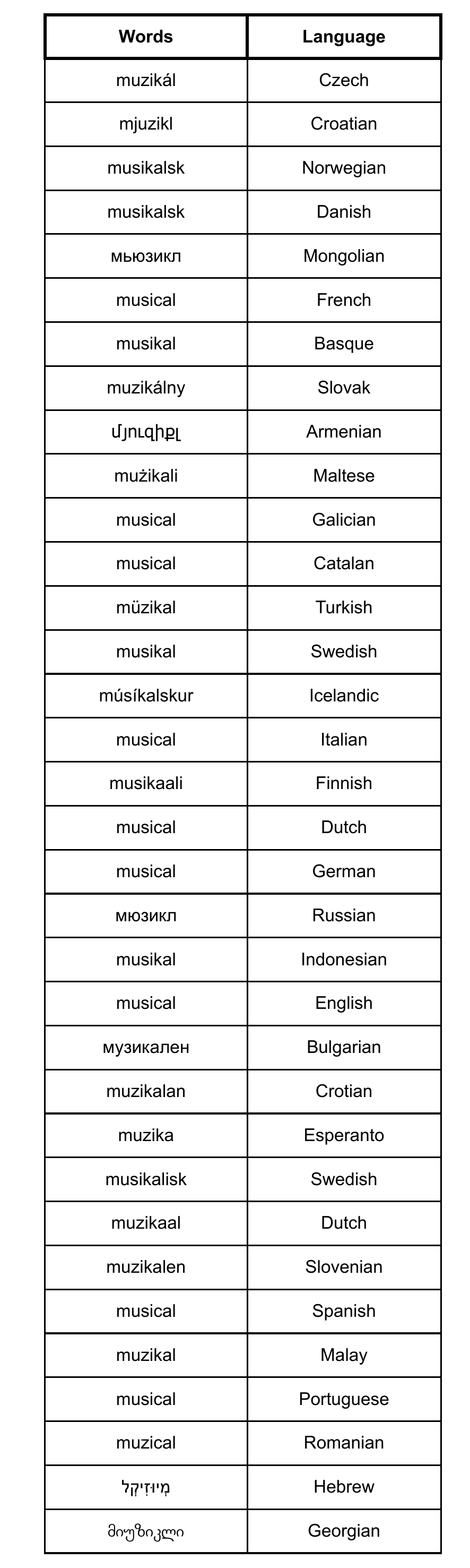}
    \caption{CogNet entries related to ``musical'' across multiple languages, including both cognates and loanwords. The dataset includes phonetically and orthographically similar words from multiple language families.}
    \label{fig:cognate-musical}
\end{figure}

Cognate words are words in different languages that share a common etymological origin, whereas loanwords are words borrowed across languages. Both can exhibit similar spellings and meanings, making them useful in linguistic studies and multilingual natural language processing.

CogNet \citep{batsuren2019cognet} is a linguistic resource that provides structured data on cognate words and loanwords across multiple languages. It is built using a combination of linguistic datasets and automated methods for identifying words with shared etymology. The dataset includes words from a wide variety of language families, offering a valuable resource for comparative linguistics and language modeling. In this study, we used CogNet as a dictionary that gives related cognate/loanword forms of an input word. For instance, the English word \textit{musical} is tokenized as \texttt{[mus', ical']} by the Llama3 tokenizer, a segmentation that does not align well with its phonological structure. This misalignment may stem from the word's extensive cross-linguistic variation, as \textit{musical} has numerous related forms across different languages. Some of these, such as \textit{muzikal} in Czech and \textit{mjuzikl} in Croatian, exhibit distinct orthographic representations that may influence tokenization irregularities.

CogNet identifies cognates using a combination of phonetic similarity, orthographic resemblance, and historical linguistic data. The process typically involves phonetic matching to identify words with similar pronunciation patterns across languages, orthographic similarity to detect common roots based on spelling, and etymological analysis leveraging linguistic databases and historical texts to verify word origins. Additionally, the dataset is refined using both automated algorithms and manual linguistic validation.

Figure \ref{fig:cognate-musical} is an example of cognate words for the English word \textit{musical}, as retrieved from CogNet. The table includes translations in multiple languages, showing how the word has evolved across different linguistic groups.

\section{Evaluate Prompt For Phonology Inference}
\label{app:evaluate prompt}
In this section, we demonstrate the prompt we used to evaluate the phonology three phonology-related tasks.

\begin{figure}[!h]
    \centering
    \begin{subfigure}[b]{\columnwidth}
        \centering
        \includegraphics[width=\linewidth]{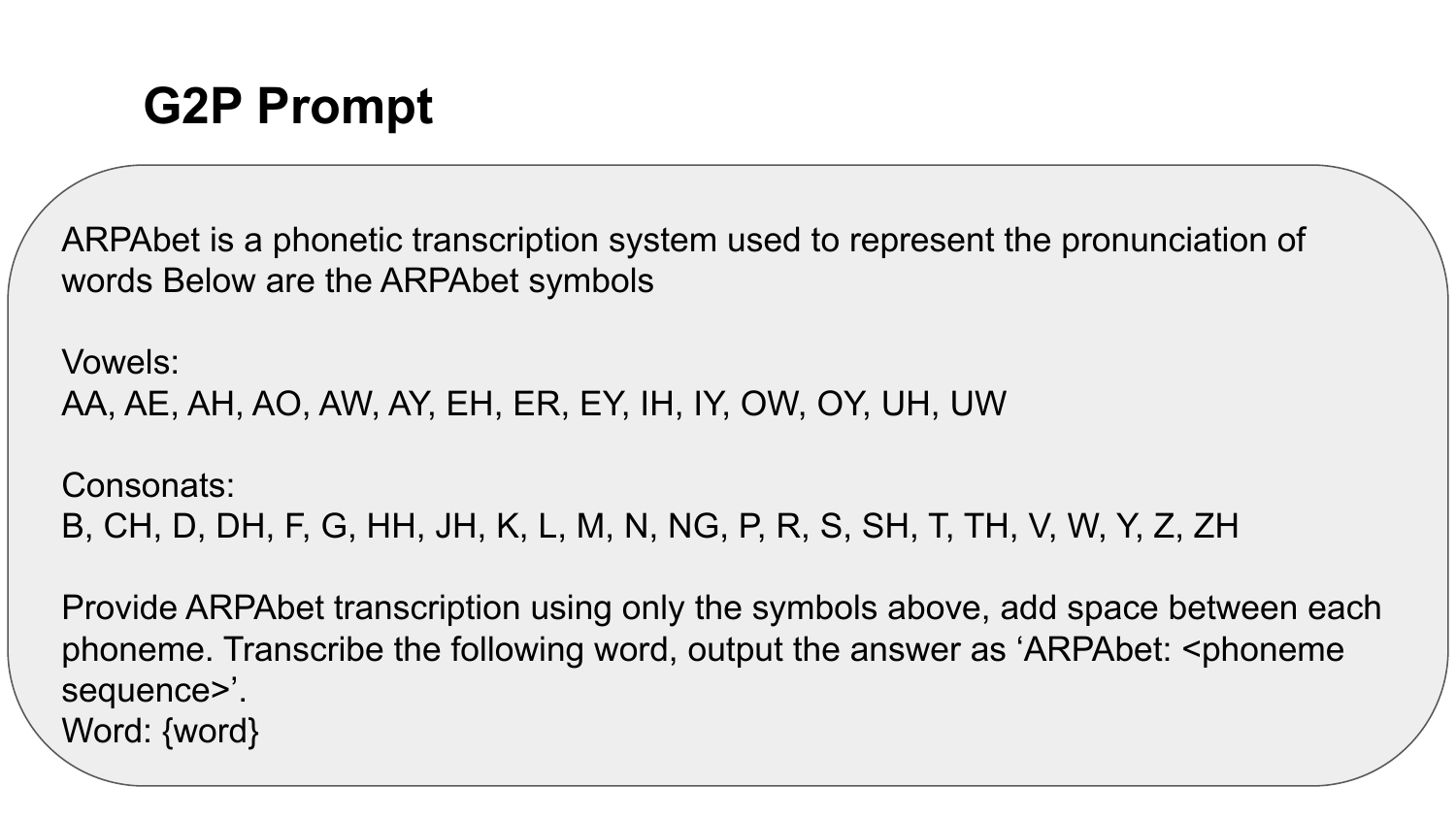}
        \caption{The prompt template we used to evaluate the G2P task.}
    \end{subfigure}

    \vspace{0.0cm}
    \begin{subfigure}[b]{\columnwidth}
        \centering
        \includegraphics[width=\linewidth]{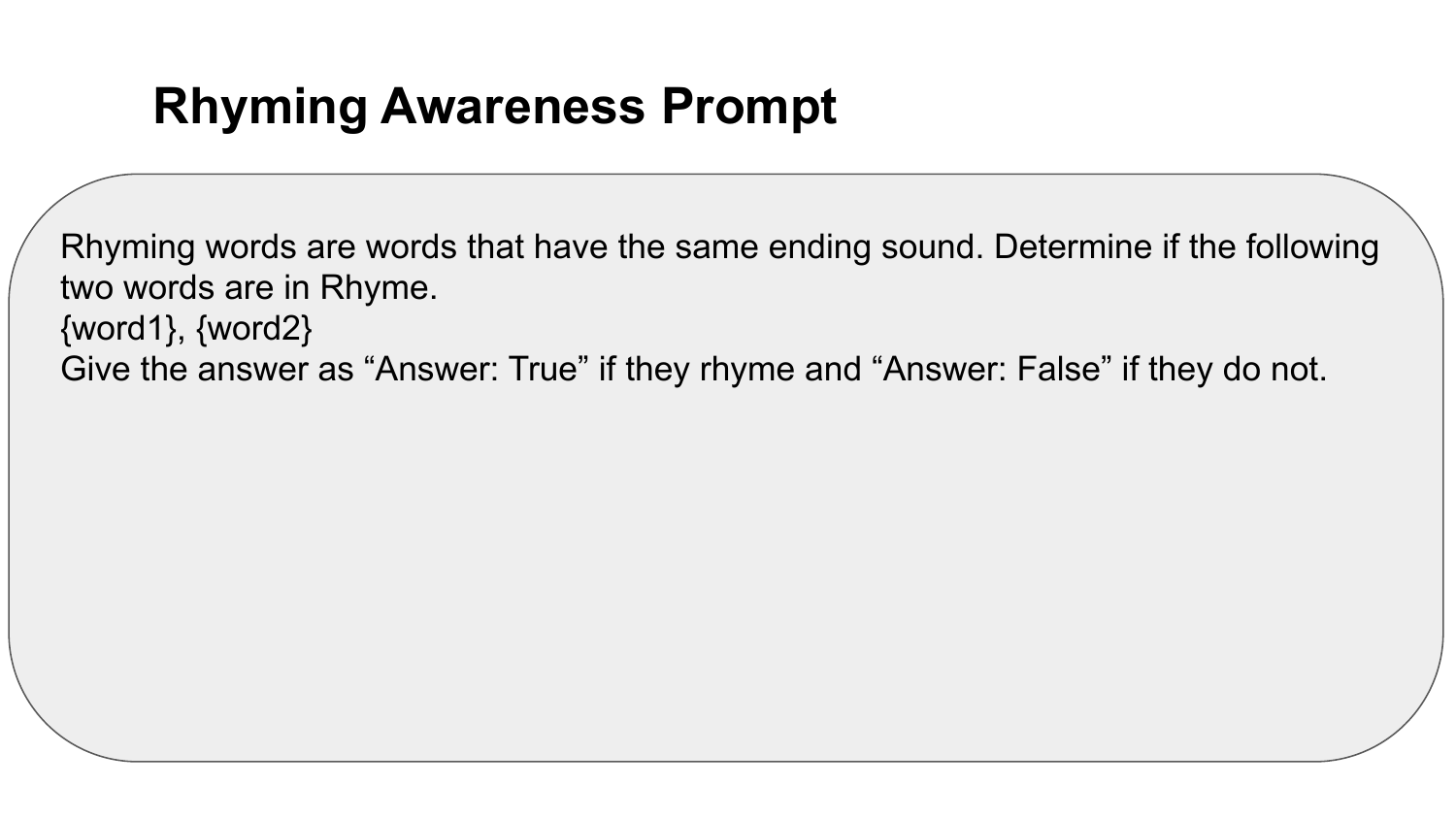}
        \caption{The prompt template we used to evaluate the rhyming awareness task.}
    \end{subfigure}

    \vspace{0.0cm}
    \begin{subfigure}[b]{\columnwidth}
        \centering
        \includegraphics[width=\linewidth]{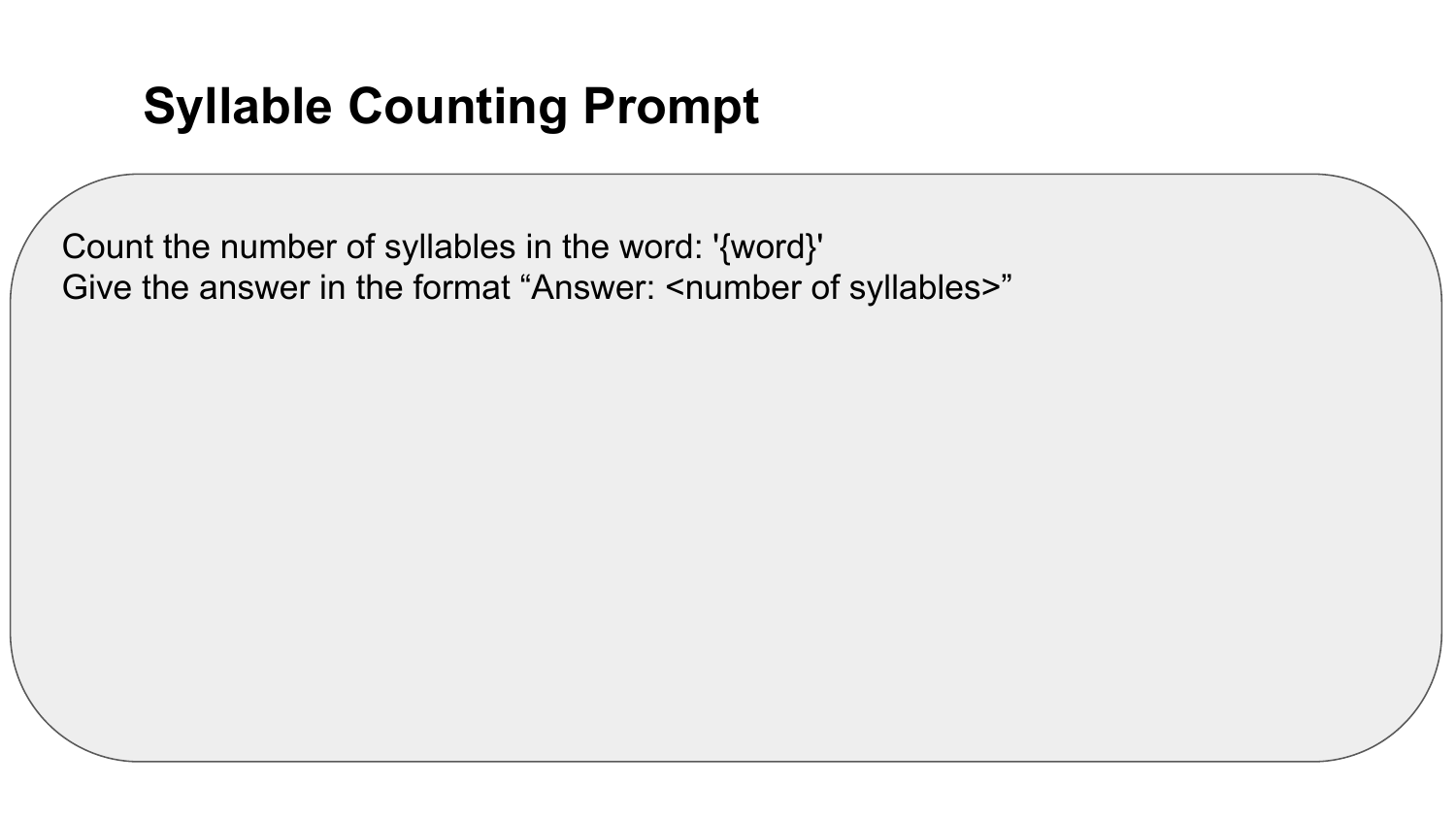}
        \caption{The prompt template we used to evaluate the syllable counting task.}
    \end{subfigure}

    \vspace{-5pt}
\end{figure}

\section{Phonology-related Task Training Template}
In \cref{alg:ipa_data_creation,alg:rhyme-creation,alg:g2p-creation,alg:syllable-creation}, \texttt{fill}$(T, x)$ denotes simple template instantiation: it replaces the placeholders in a question or answer template $T$ with the provided content $x$ (e.g., IPA transcription of sampled words).

\begin{algorithm}[!h]
    \footnotesize
    \caption{Conversation Data Creation with IPA Annotations}
    \label{alg:ipa_data_creation}
    \begin{algorithmic}[1]
        \Require Dataset $\mathcal{D}$ with question-answer pairs $(q, a)$, IPA sentence template $L$. Function \texttt{fill}$(T, w)$ to fill a template $T$ using word $w$. Function \texttt{get\_IPA} to get IPA.
        \Ensure Modified dataset $\mathcal{D'}$ with IPA-annotated questions and answers

        \State $\mathcal{D'} \gets \emptyset$
        \For{each $(q, a) \in \mathcal{D}$}
        \State Split $q$ into words: $W \gets \text{split}(q)$
        \State $q' \gets q$
        \State Sample $k \sim \text{Uniform}(\{0,1,2\})$
        \State Uniformly sample $S \subset W$ with $|S| = k$
        \State $S_{\text{IPA}} \gets \emptyset$
        \For{each selected word $w_i \in S$}
        \State Replace $w_i$ in $q'$ with <IPA>$w_i$</IPA>.
        \State Obtain IPA transcription of $w_i$: $p_i = \text{get\_IPA}(w_i)$
        \State Add $(w_i, p_i)$ to $S_{\text{IPA}}$
        \EndFor
        \State filling in words from $S_{\text{IPA}}$ into $L$, $l = \text{fill}(L, S_{\text{IPA}})$
        \State Prepend sentence $l$ indicating IPA representation to $a$: $a' \gets l + a$
        \State Add modified pair $(q', a')$ to $\mathcal{D'}$
        \EndFor
        \Return $\mathcal{D'}$
    \end{algorithmic}
\end{algorithm}

\begin{algorithm}[!h]
    \footnotesize
    \caption{Dataset Creation for Rhyming Awareness Task}
    \label{alg:rhyme-creation}
    \begin{algorithmic}[1]
        \Require Word pair list with IPA transcriptions, possible templates $P_1, \dots, P_5$. Positive answer template $A_P$, negative answer template $A_N$. Function \texttt{fill}$(T, w)$ to fill a template $T$ using word $w$.
        \Ensure Dataset with question-answer pairs
        \For{each $(\text{word}_1, \text{word}_2)$ pair}
        \State Sample $i \sim \text{Uniform}(\{1, 2, 3, 4, 5\})$
        \State $P \leftarrow$ fill$(P_i, (\text{word}_1, \text{word}_2))$
        \State Extract IPA endings of $\text{word}_1$ and $\text{word}_2$
        \If{IPA endings match}
        \State response $\leftarrow$ fill$(A_P, (\text{word}_1, \text{word}_2))$
        \Else
        \State response $\leftarrow$ fill$(A_N, (\text{word}_1, \text{word}_2))$
        \EndIf
        \State Store $(P, \text{response})$ in dataset
        \EndFor
    \end{algorithmic}
\end{algorithm}

\begin{algorithm}[!h]
    \footnotesize
    \caption{Dataset Creation for Grapheme-to-Phoneme (G2P) Task}
    \label{alg:g2p-creation}
    \begin{algorithmic}[1]
        \Require Word list with IPA transcriptions, possible templates $P_1, \dots, P_5$. Answer template $A$. ARPAbet-to-phoneme dictionary $M$. Function \texttt{fill}$(T, w)$ to fill a template $T$ using word $w$.
        \Ensure Dataset with question-answer pairs
        \For{each word $w$ in dataset}
        \State Sample $i \sim \text{Uniform}(\{1, 2, 3, 4, 5\})$
        \State $P \leftarrow$ fill$(P_i, w)$
        \State Obtain IPA transcription of $w$: $I = \text{get\_IPA}(w)$
        \State IPA phonemes to ARPAbet: $A = [M[p] \text{ for } p \in I]$
        \State response $\leftarrow$ fill$(A, (w, I, A))$
        \State Store $(P, \text{response})$ in dataset
        \EndFor
    \end{algorithmic}
\end{algorithm}

\begin{algorithm}[!h]
    \footnotesize
    \caption{Dataset Creation for Syllable Counting Task}
    \label{alg:syllable-creation}
    \begin{algorithmic}[1]
        \Require Word list with IPA transcriptions, possible templates $P_1, \dots, P_5$. Answer template $A$. Function \texttt{fill}$(T, w)$ to fill a template $T$ using word $w$.
        \Ensure Dataset with question-answer pairs
        \For{each word $w$ in dataset}
        \State Sample $i \sim \text{Uniform}(\{1, 2, 3, 4, 5\})$
        \State $P \leftarrow$ fill$(P_i, w)$
        \State Obtain IPA transcription of $w$: $I = \text{get\_IPA}(w)$
        \State Identify vowel nuclei (monophthongs or diphthongs) in $I$
        \State Compute syllable count: $S = \text{count\_syllables}(I)$
        \State Format response using $S$: response $\leftarrow$ fill$(A, (w, S))$
        \State Store $(P, \text{response})$ in dataset
        \EndFor
    \end{algorithmic}
\end{algorithm}

\label{app:data example}
Here, we demonstrate the detailed training template we used for constructing the training dataset for each problem. We demonstrate how we construct 4 categories of QA pairs as our fine-tuning dataset.

\begin{figure}[!h]
    \centering
    \begin{subfigure}[b]{\columnwidth}
        \centering
        \includegraphics[width=\linewidth]{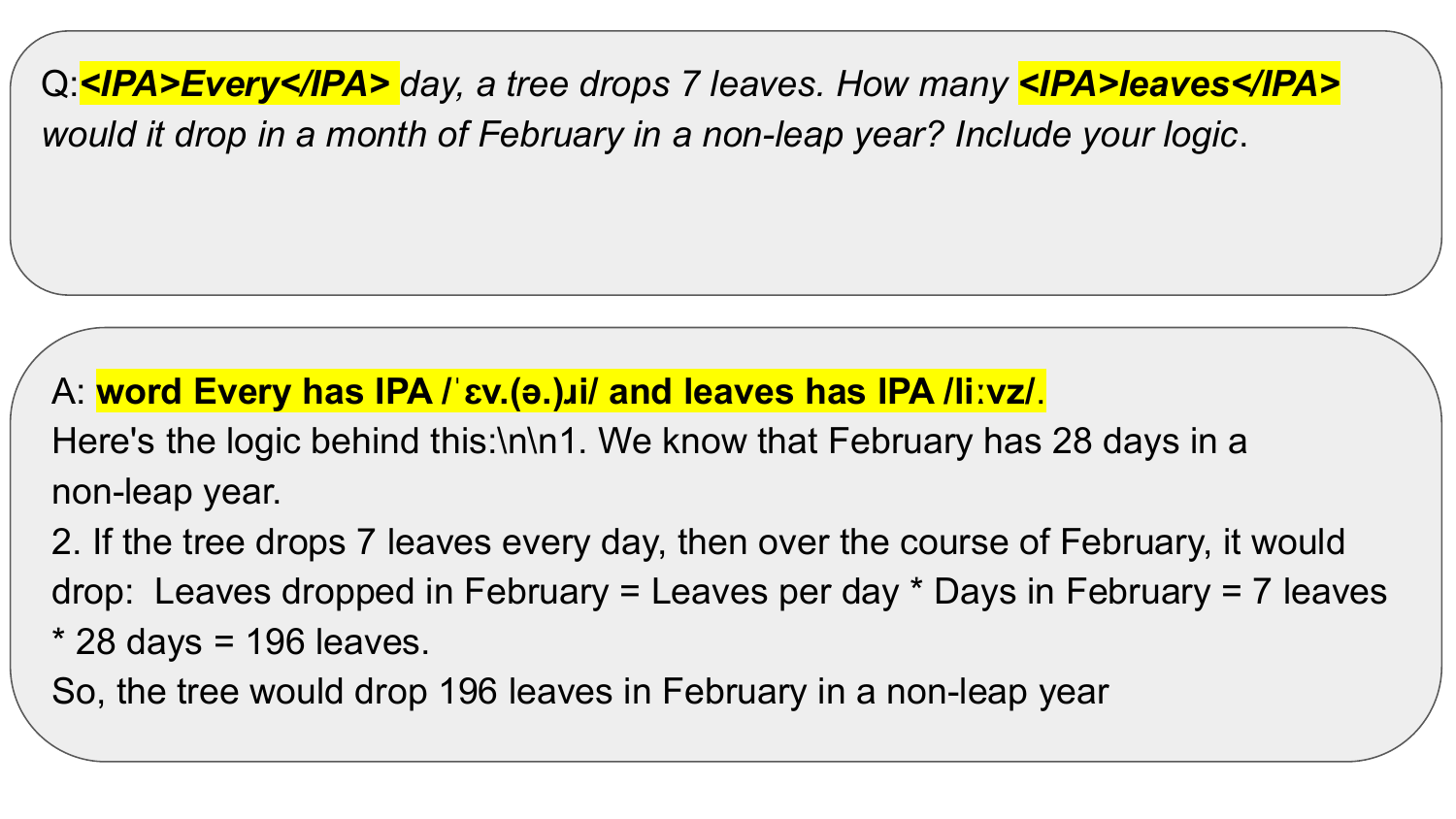}
        \caption{An example of how we construct QA data from OpenHermes2.5.}
        \label{fig:conversation-example}
    \end{subfigure}

    \vspace{0.0cm}
    \begin{subfigure}[b]{\columnwidth}
        \centering
        \includegraphics[width=\linewidth]{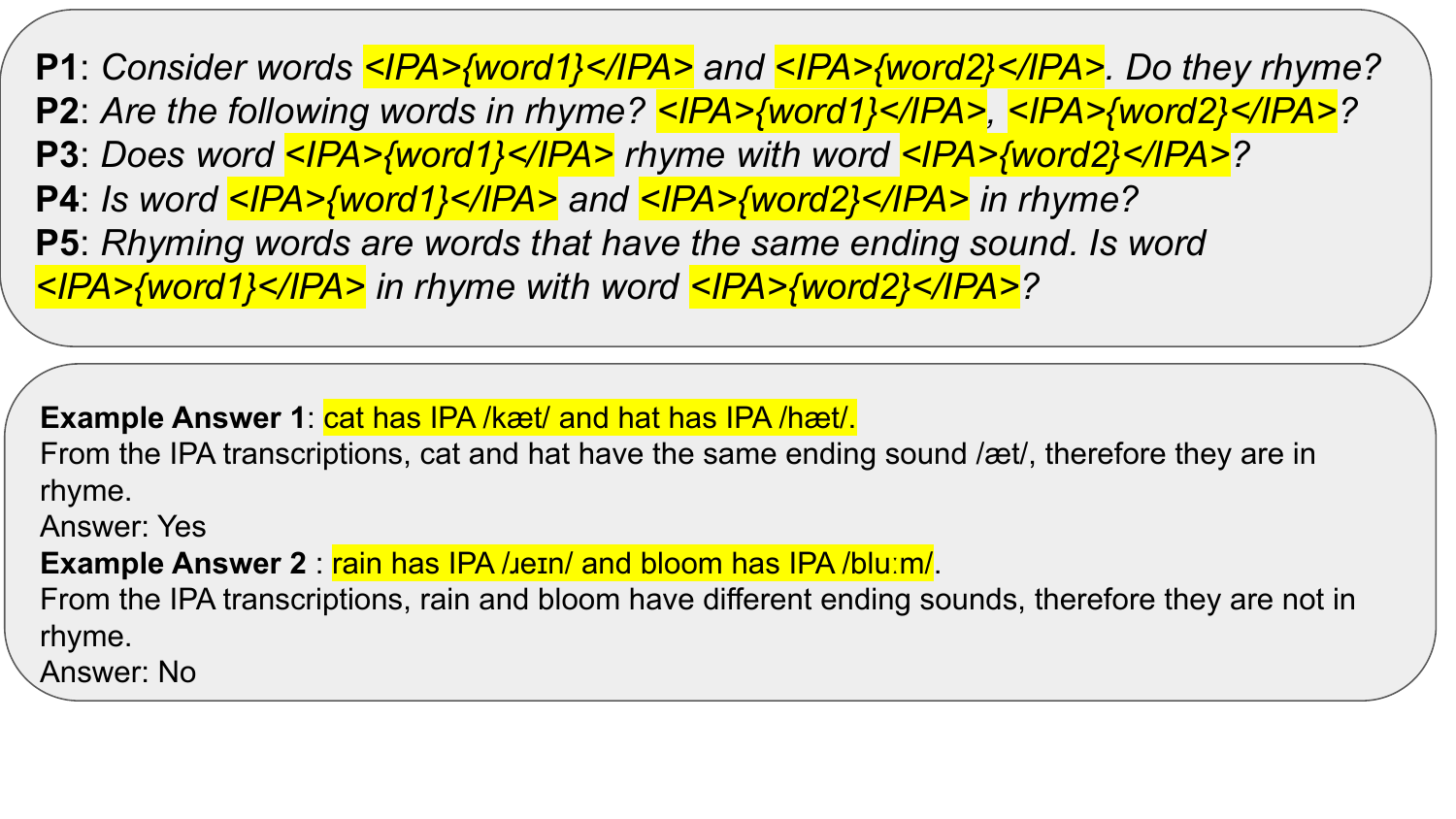}
        \caption{All possible questions and example answers for the rhyming awareness task.}
        \label{fig:rhyme-train}
    \end{subfigure}

    \vspace{0.0cm}
    \begin{subfigure}[b]{\columnwidth}
        \centering
        \includegraphics[width=\linewidth]{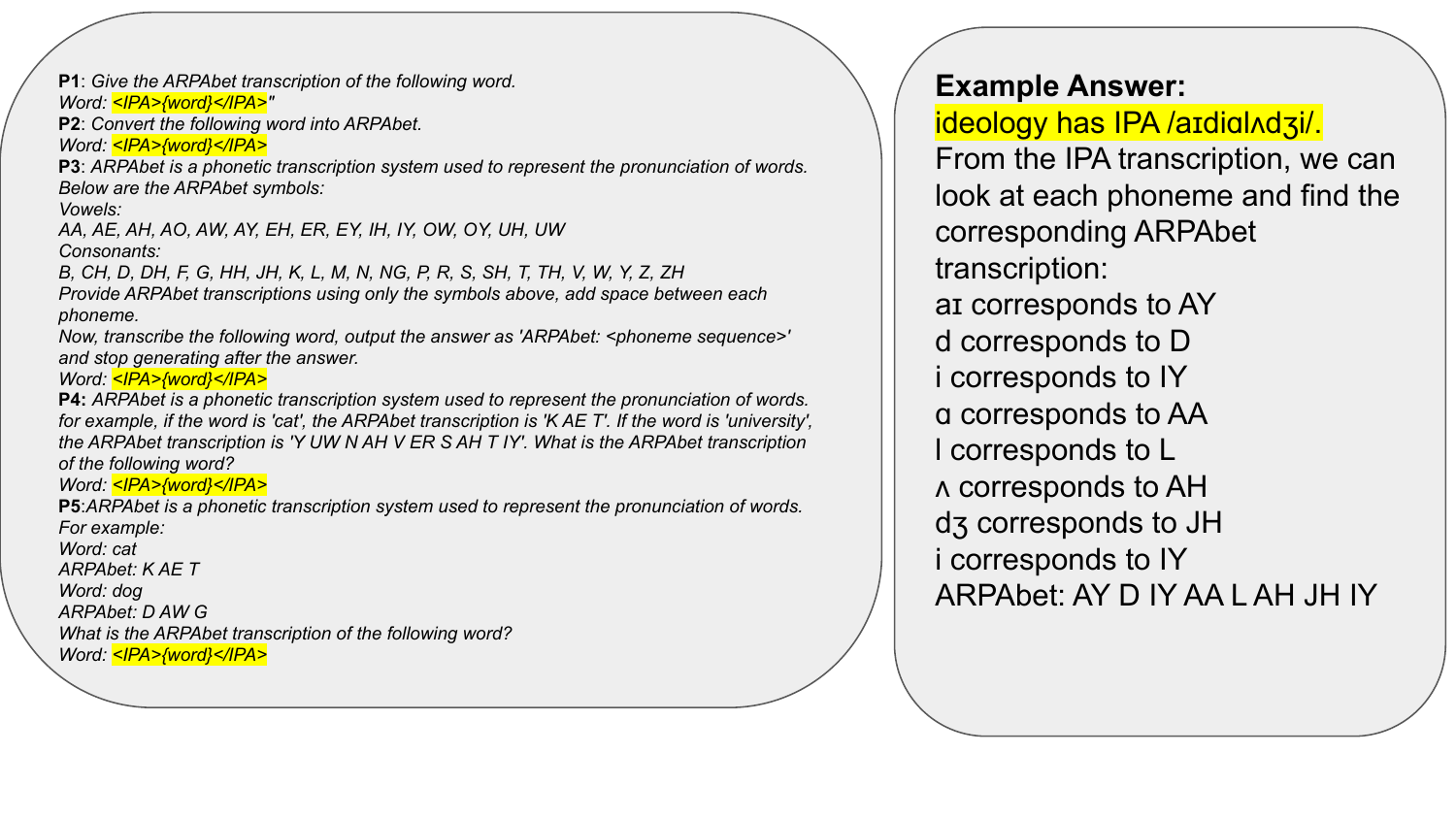}
        \caption{All possible questions and an example answer for G2P task.}
        \label{fig:g2p-train}
    \end{subfigure}

    \vspace{0.0cm}
    \begin{subfigure}[b]{\columnwidth}
        \centering
        \includegraphics[width=\linewidth]{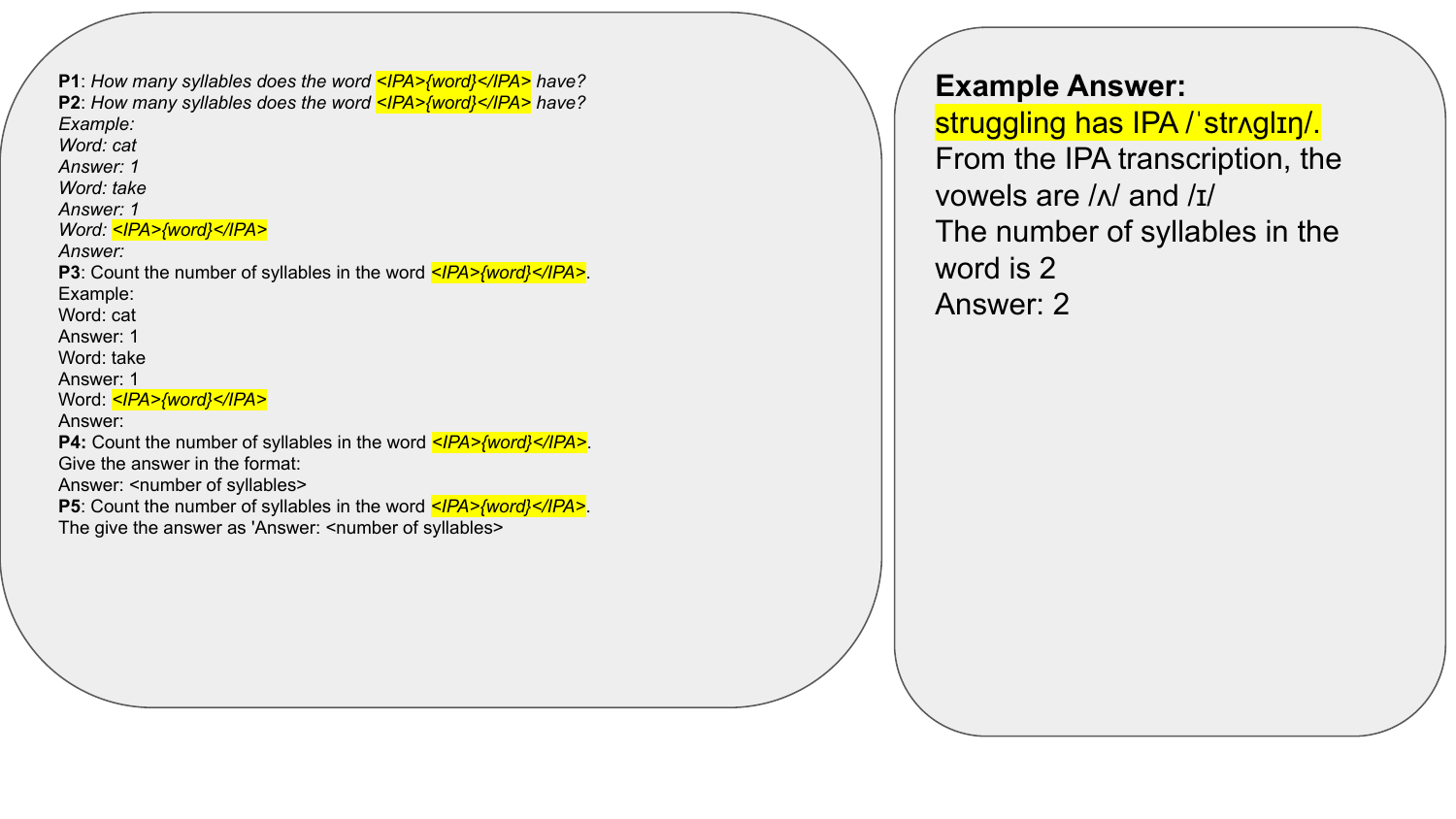}
        \caption{All possible questions and an example answer for syllable counting task.}
        \label{fig:syllable-train}
    \end{subfigure}
    \caption{Examples of our question templates and some example answers. The \hl{yellow part} is the common part of the fine-tuning dataset, which helps the model to identify which word to consider IPA and give the IPA explicitly.}
    \vspace{-5pt}
\end{figure}

\noindent\vspace{2pt}\textbf{Conversation.} We used OpenHermes2.5 \citep{OpenHermes2.5} as the source of the conversation dataset, it involves all kinds of conversational datasets consisting of question and answer. In the question, we randomly select 0 - 2 words and wrap the words with an IPA token to indicate that we want the IPA of the word, and in the answer, we add one sentence indicating the IPA of the words. This augmentation is intentionally orthogonal to the QA semantics: it preserves the original task while training the model to attend to phonological side information. We present the process of data creation in \cref{alg:ipa_data_creation} and show an example in \cref{fig:conversation-example}.

\noindent\vspace{2pt}\textbf{Rhyming Awareness.} In the rhyming awareness, we prepare 5 possible question templates $P_1, P_2, \dots P_5$ to mimic the possible users' questions. In the answer, we first give the IPA of the word as in Conversation. Then, from the IPA, we extract the same part of the IPA if two words are in rhyme, or state two words are not in rhyme if the IPA does not have the same ending. We present the process of rhyming awareness data creation in \cref{alg:rhyme-creation} and we show the question templates and example answer in \cref{fig:rhyme-train}.

\noindent\vspace{2pt}\textbf{G2P.} In the G2P task, we also prepare 5 possible question templates. In the answer, we break the IPA transcript phoneme by phoneme and use the map from phoneme to ARPAbet to convert the IPA to ARPAbet. We present the process of G2P data creation in \Cref{alg:g2p-creation} and the question templates and an example answer in \Cref{fig:g2p-train}.

\noindent\vspace{2pt}\textbf{Syllable Counting.} In the syllable counting task, there are also 5 possible questions as before. In the solution, we use a simple English IPA heuristic: each vowel nucleus (a monophthong or diphthong) contributes one syllable, so counting these nuclei gives the total number of syllables. We present the process of syllable counting data creation in \cref{alg:syllable-creation} and the question templates and an example answer in \cref{fig:syllable-train}.
\section{Controlled-Experiment of the Probing}
\label{app:control-exp}

\begin{figure*}[!ht]
    \centering
    \includegraphics[width=\linewidth]{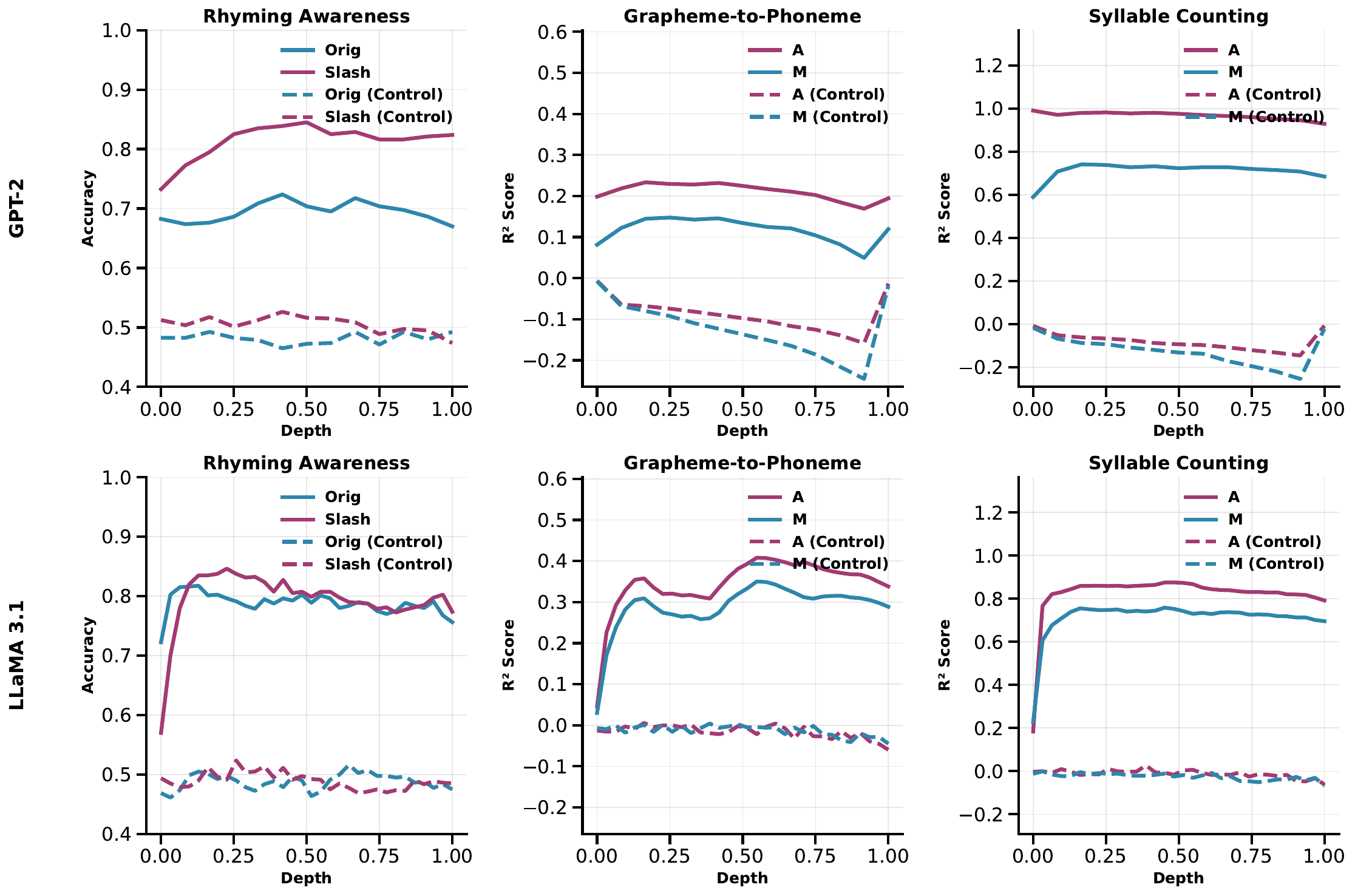}
    \caption{\textbf{Control-label sanity check.}
        Probing performance with \emph{random} targets for GPT-2 (upper block) and Llama-3.1-8B (lower block).
        Left-to-right: (i) Rhyming awareness--accuracy; (ii) G2P--\(R^{2}\); (iii) syllable counting--\(R^{2}\).
        Solid lines reproduce the original probes, dashed lines the corresponding control probes.
        All curves collapse to chance (accuracy $\approx$ 0.5) or sub-chance (\(R^{2}\le 0\)), demonstrating that the linear probes do not overfit when the target carries no linguistic signal.}
    \label{fig:control-experiment}
\end{figure*}

To verify that our linear probes do not artificially inflate performance, we repeat every probing experiment with \emph{randomly generated targets}.
For rhyming awareness, we assign a random binary label to each word-pair; for grapheme-to-phoneme (G2P) prediction we draw a random integer in the range \([0,39]\) for every phoneme slot; and for syllable counting, we sample a random integer between 0 and 8.
We train the same logistic- and ridge-regression probes as in the main study on GPT-2 and Llama-3.1-8B using these synthetic labels.

As summarized in \cref{fig:control-experiment}, the control probes behave exactly as expected: accuracy hovers around the chance rate of 0.5 for the binary classification task, and all \(R^{2}\) values for the two regression tasks are zero or negative across layers.
This confirms that the probes themselves lack the capacity to memorize arbitrary labelings and that the positive results reported in the main paper genuinely stem from information encoded in the models’ hidden representations rather than from overfitting artifacts.

\section{Probing Details}
\label{app:probing-details}
If we have $n$ words/pairs of words as input, after prompting them to LMs, for each layer $l$, we will get a matrix of $\mH_l \in \mathbb{R}^{n \times d}$, where $d$ is the dimension of the hidden states. Then, if we have the ground truth $\vy$, we can train models using $\mH_l$ and $\vy$, and we will discuss the details of our probing for each task. For the model we trained, we used the scikit-learn \cite{scikit-learn} implementation. For each experiment, we ran 10 times with seeds 0 - 9, and did an 80 - 20 train-test split, reporting the metrics on the test set. We only selected a linear model for evaluation, since the goal of our work is not to achieve high performance on the downstream task but
to illustrate the bias introduced by the tokenizers. Also, \citet{hewitt2019designing} illustrated that using a complex model like a Neural Network may cause the probing result to be unreliable since the model will learn the feature, and our evaluated tasks are not very hard tasks, thus, the linear model is enough to reveal the representation quality of different words. To obtain the embeddings, we use a single A40 GPU and do a single forward pass.

\noindent\vspace{2pt}\textbf{Rhyming Awareness}. For the rhyming awareness task, the input is a pair of words, and ground truth $\vy \in \sR^{n}$ is a binary label indicating if the words pair is rhyming. Then we used logistic regression \texttt{LogisticRegression}, and we set the max iterations to 1000, and Inverse of regularization strength $C = 10$, other hyperparameters are set as default. We trained two logistic regression classifiers on both original words and words with slash inserted.

\noindent\vspace{2pt}\textbf{G2P.} For the G2P task, the label is the categorical encoding of the ARPAbet symbols (0 - 39), we either truncated or padded the label with 0. Therefore, we have $\mY \in \sR^{n \times 8}$. We used the Cross-validation Ridge regression \texttt{RidgeCV} to regress the label and set the alphas to be chosen from $\{10, 100, 500, 1000, 2000\}$, other hyperparameters are set as default. We train two ridge regressors on both syllable-token aligned and misaligned groups.

\noindent\vspace{2pt}\textbf{Syllable Counting.} For the syllable counting task, the label is the number of syllables in the word. Therefore, we have $\vy \in \sR^{n}$. We also used Cross-validation Ridge regression \texttt{RidgeCV} to regress the label and set alphas to be chosen from $\{10, 100, 500, 1000, 2000\}$, other hyperparameters are set as default. We train two ridge regressors on both syllable-token aligned and misaligned groups.

\section{Fine-tuning \& Evaluation Details}
\label{app:finetune-details}

For the evaluation, we used the chat template of the corresponding model to form the QA. And we used vllm \citep{kwon2023efficient} and set the decoding strategy with \texttt{top-p}=0.95 and \texttt{temperature}=0.8.

We do not report task-specific non-neural baselines in the main performance-based evaluation because rhyming awareness and syllable counting labels are derived from CMU-based dictionary/rule preprocessing, making lookup-based systems near-oracle label generators, whereas dedicated G2P systems are separately supervised models and are therefore not directly comparable to zero-shot evaluation of pretrained foundation models.

For the fine-tuning, we leveraged the Hugging Face \texttt{transformers} library alongside Parameter-Efficient Fine-Tuning (PEFT) to integrate LoRA \citep{hu2021loralowrankadaptationlarge}. We specifically targeted the query (q\_proj) and value (v\_proj) projection layers for adaptation. We set the LoRA Rank $(r)$ to 8, LoRA scaling factor $(\alpha)$ to 16, LoRA Dropout to 0.1.

For multi-GPU training, we employed Hugging Face Accelerate, which facilitated seamless distributed training across the two A40 GPUs using Pytorch Distrubuted Data Parallel (DDP) for less than 5 GPU hours. The model and dataset were automatically partitioned and synchronized, ensuring efficient computation.

\section{Rhyming probing using Different Delimiters}
\label{app:delimiter}
\definecolor{mygray}{gray}{0.85}  
\begin{table}
  \setlength{\tabcolsep}{3pt}
  \footnotesize
  \centering
  \begin{tabular}{cc|cccccc}
    \toprule
    \textbf{Model} & \textbf{Delimiter} &
    \multicolumn{6}{c}{\textbf{Depth} (Accuracy $\uparrow$)}                       \\
    \cmidrule(lr){3-8}
    \textbf{Size}  &                    & 0\%  & 20\% & 40\% & 60\% & 80\% & 100\% \\
    \midrule
    \multicolumn{8}{c}{\textit{Sub-word Tokenization}}                             \\
    \midrule
    \rowcolor{mygray}
    \multicolumn{8}{l}{\textbf{BERT}}                                         \\
    110M           & None               & 56.0 & 67.6 & 68.3 & 70.9 & 71.0 & 70.5  \\
                   & Slash              & 68.6 & 74.5 & 73.4 & 77.5 & 79.5 & 78.1  \\
                   & Comma              & 68.6 & 75.7 & 71.8 & 77.8 & 80.3 & 79.2  \\
                   & Dot                & 68.6 & 74.7 & 73.6 & 80.8 & 79.8 & 78.4  \\
    \midrule
    \rowcolor{mygray}
    \multicolumn{8}{l}{\textbf{GPT-2}}                                        \\
    124M           & None               & 63.4 & 64.7 & 66.1 & 66.2 & 66.0 & 61.6  \\
                   & Slash              & 71.7 & 76.9 & 77.2 & 79.1 & 78.5 & 77.5  \\
                   & Comma              & 71.7 & 77.6 & 77.1 & 78.8 & 77.9 & 77.3  \\
                   & Dot                & 72.3 & 77.3 & 77.6 & 79.5 & 78.4 & 76.9  \\
    \midrule
    \rowcolor{mygray}
    \multicolumn{8}{l}{\textbf{Llama-3.1-8B-Instruct}}                             \\
    8B             & None               & 72.5 & 79.8 & 79.0 & 77.9 & 77.3 & 74.9  \\
                   & Slash              & 56.3 & 85.1 & 84.0 & 80.0 & 78.9 & 79.5  \\
                   & Comma              & 56.8 & 86.7 & 83.7 & 79.3 & 78.5 & 77.4  \\
                   & Dot                & 56.7 & 85.2 & 82.0 & 79.8 & 77.8 & 76.3  \\
    \midrule
    \rowcolor{mygray}
    \multicolumn{8}{l}{\textbf{Mistral-7B-Instruct-v3}}                            \\
    7B             & None               & 64.5 & 80.6 & 80.8 & 78.8 & 77.4 & 74.7  \\
                   & Slash              & 55.8 & 81.1 & 82.7 & 79.5 & 79.0 & 77.6  \\
                   & Comma              & 55.8 & 81.7 & 85.4 & 82.1 & 80.3 & 79.5  \\
                   & Dot                & 55.8 & 80.5 & 81.7 & 79.1 & 77.9 & 77.2  \\
    \bottomrule
  \end{tabular}
  \caption{Ablation study of delimiter formats (“None”, “Slash”, “Comma”, “Dot”) across different depths of the hidden states.}
  \label{tab:delimiter-ablation}

  \vspace{-5mm}
\end{table}

In the rhyming awareness probing experiment, we initially used the slash (``/'') as a delimiter to split word pairs, enabling more structured and representative hidden states. To assess the robustness of this delimiter choice—and to test whether performance gains stem from improved tokenization granularity rather than the specific symbol—we conducted an ablation study using alternative delimiters: the comma (``,'') and the dot (``.'').

We evaluated probing performance across four language models—BERT, GPT-2, LLaMA3.1-8B, and Mistral-7B—and report results in \cref{tab:delimiter-ablation}. Across all models, the probers trained with any delimiter (slash, comma, or dot) yield comparable performance throughout the depth of the hidden layers. Importantly, all delimiter-based variants consistently outperform the baseline where no delimiter is used (None), confirming that the performance gains are primarily due to the introduction of fine-grained structure in the input rather than the specific choice of delimiter.
\begin{table*}[h]
    \setlength{\tabcolsep}{3.4pt}
    \renewcommand{\arraystretch}{0.95}
    \footnotesize
    \centering
    \begin{tabular}{ccrrrrrrrrrrrrrr}
        \toprule
        \multirow{2}{*}{\textbf{Model}}                                      & \multirow{2}{*}{\textbf{STAD}} & \hspace{-2pt} & \multicolumn{6}{c}{\textbf{G2P} ($R^2$ by Layer Depth)} & \hspace{-2pt}    & \multicolumn{6}{c}{\textbf{Syllable Counting} ($R^2$ by Layer Depth)}                                                                                                                                                                                                        \\
        \cmidrule(lr){4-9} \cmidrule(lr){11-16}
                                                                             & \hspace{-2pt}                  &               & 0\%                                                     & 20\%             & 40\%                                                                  & 60\%               & 80\%             & 100\%            & \hspace{-2pt} & 0\%                & 20\%               & 40\%               & 60\%               & 80\%               & 100\%            \\
        \cmidrule(lr){1-2} \cmidrule(lr){4-9} \cmidrule(lr){11-16}
        \cellcolor{mygray}                                                   & .000 (A)                       & \hspace{-2pt} & .004                                                    & .056             & .001                                                                  & .002               & \textbf{.030}    & .009             & \hspace{-2pt} & {\threestar{.999}} & \threestar{.952}   & .009               & .022               & .129               & .054             \\
        \multirow{-2}{*}{\cellcolor{mygray} BERT}                            & .290 (M)                       & \hspace{-2pt} & \bf .009                                                & \bf.085          & .001                                                                  & .002               & .024             & \bf .010         & \hspace{-2pt} & .404               & .626               & \bf .068           & \bf .074           & \bf .264           & \bf .143         \\
        \midrule
        \cellcolor{mygray}                                                   & .000 (A)                       & \hspace{-2pt} & \threestar{.198}                                        & \threestar{.229} & {\threestar{.232}}                                                    & \threestar{.217}   & \threestar{.185} & \threestar{.194} & \hspace{-2pt} & .027               & {\threestar{.980}} & \threestar{.980}   & \threestar{.969}   & \threestar{.952}   & \threestar{.929} \\
        \multirow{-2}{*}{\cellcolor{mygray} GPT-2}                           & .388 (M)                       & \hspace{-2pt} & .081                                                    & {.148}           & .146                                                                  & .124               & .080             & .119             & \hspace{-2pt} & \bf .589           & {.740}             & .732               & .728               & .714               & .684             \\
        \midrule
        \cellcolor{mygray}                                                   & .000 (A)                       & \hspace{-2pt} & .058                                                    & \twostar{.238}   & \onestar{.231}                                                        & \textbf{.222}      & \textbf{.214}    & \twostar{.193}   & \hspace{-2pt} & .476               & \threestar{.947}   & {\threestar{.966}} & \threestar{.950}   & \threestar{.936}   & \threestar{.922} \\
        \multirow{-2}{*}{\cellcolor{mygray} BLOOM-560M}                      & .376 (M)                       & \hspace{-2pt} & \bf.096                                                 & .196             & .215                                                                  & .215               & .199             & .168             & \hspace{-2pt} & \bf .489           & .766               & .753               & .711               & .674               & .608             \\
        \midrule
        \cellcolor{mygray}                                                   & .000 (A)                       & \hspace{-2pt} & \threestar{.179}                                        & {\twostar{.219}} & \threestar{.211}                                                      & \threestar{.148}   & \threestar{.078} & \threestar{.004} & \hspace{-2pt} & \threestar{.945}   & \threestar{.953}   & {\threestar{.967}} & \threestar{.942}   & \threestar{.930}   & \threestar{.914} \\
        \multirow{-2}{*}{\cellcolor{mygray} GPT-neo-2.7b}                    & .388 (M)                       & \hspace{-2pt} & .072                                                    & {.169}           & .111                                                                  & .005               & -.124            & -.219            & \hspace{-2pt} & .555               & {.787}             & .758               & .692               & .634               & .539             \\
        \midrule
        \cellcolor{mygray}                                                   & .000 (A)                       & \hspace{-2pt} & \textbf{.001}                                           & .212             & {\textbf{.301}}                                                       & .314               & \textbf{.297}    & .239             & \hspace{-2pt} & .028               & \threestar{.800}   & {\threestar{.913}} & \threestar{.911}   & \threestar{.854}   & \threestar{.816} \\
        \multirow{-2}{*}{\cellcolor{mygray} Mistral-7b-Instruct-v3}          & .348 (M)                       & \hspace{-2pt} & .000                                                    & \bf .214         & .283                                                                  & \bf {.317}         & .282             & \bf .261         & \hspace{-2pt} & \bf .045           & .708               & .804               & {.806}             & .789               & .762             \\
        \midrule
        \cellcolor{mygray}                                                   & .000 (A)                       & \hspace{-2pt} & \threestar{.168}                                        & \onestar{.279}   & \textbf{.247}                                                         & \textbf{.260}      & {\onestar{.312}} & \textbf{.193}    & \hspace{-2pt} & .383               & \threestar{.921}   & \threestar{.934}   & {\threestar{.976}} & \threestar{.946}   & \onestar{.782}   \\
        \multirow{-2}{*}{\cellcolor{mygray} gemma-7b}                        & .303 (M)                       & \hspace{-2pt} & .054                                                    & .229             & .231                                                                  & .226               & {.284}           & .183             & \hspace{-2pt} & \bf.640            & {.773}             & .769               & .772               & .758               & .672             \\
        \midrule
        \cellcolor{mygray}                                                   & .000 (A)                       & \hspace{-2pt} & \threestar{.034}                                        & \threestar{.349} & \threestar{.356}                                                      & {\threestar{.370}} & \threestar{.366} & \threestar{.325} & \hspace{-2pt} & .152               & \threestar{.931}   & {\threestar{.935}} & \threestar{.923}   & \threestar{.899}   & \threestar{.860} \\
        \multirow{-2}{*}{\cellcolor{mygray} Llama3-8b-Instruct}              & .372 (M)                       & \hspace{-2pt} & .023                                                    & .295             & .297                                                                  & {.333}             & .308             & .276             & \hspace{-2pt} & \bf.165            & .769               & {.795}             & .769               & .749               & .717             \\
        \midrule
        \cellcolor{mygray}                                                   & .000 (A)                       & \hspace{-2pt} & \textbf{.033}                                           & \onestar{.325}   & \twostar{.321}                                                        & {\twostar{.387}}   & \twostar{.357}   & \textbf{.166}    & \hspace{-2pt} & .188               & \threestar{.936}   & {\threestar{.939}} & \threestar{.921}   & \threestar{.898}   & \threestar{.859} \\
        \multirow{-2}{*}{\cellcolor{mygray} Llama3.1-8b-Instruct}            & .372 (M)                       & \hspace{-2pt} & .029                                                    & .304             & .285                                                                  & {.349}             & .317             & .157             & \hspace{-2pt} & \bf.211            & .783               & {.789}             & .769               & .754               & .723             \\
        \midrule

        \cellcolor{mygray}                                                   & .000 (A)                       & \hspace{-2pt} & \threestar{.100}                                        & \twostar{.209}   & {\twostar{.192}}                                                      & \threestar{.153}   & \threestar{.151} & \threestar{.149} & \hspace{-2pt} & .419               & \threestar{.974}   & {\threestar{.977}} & \threestar{.975}   & \threestar{.975}   & \threestar{.977} \\
        \multirow{-2}{*}{\cellcolor{mygray} Falcon3-7b-Instruct}             & .337 (M)                       & \hspace{-2pt} & .050                                                    & {.148}           & .155                                                                  & .054               & .004             & .025             & \hspace{-2pt} & \bf .618           & {.776}             & .769               & .734               & .728               & .729             \\
        \midrule
        \cellcolor{mygray}                                                   & .000 (A)                       & \hspace{-2pt} & .056                                                    & \textbf{.240}    & {\textbf{.269}}                                                       & \onestar{.245}     & \twostar{.210}   & \twostar{.189}   & \hspace{-2pt} & .252               & \twostar{.925}     & \threestar{.937}   & \threestar{.940}   & {\threestar{.941}} & \threestar{.936} \\
        \midrule
        \multicolumn{2}{l}{\cellcolor{mygray} Control: Randomized Embedding} & \hspace{-2pt}                  & -.070         & -.101                                                   & -.043            & -.073                                                                 & -.066              & -.082            & \hspace{-2pt}    & -.082         & -.073              & .001               & -.115              & -.097              & -.022                                 \\
        \bottomrule
    \end{tabular}
    \caption{
        $R^2$ for G2P and syllable counting, comparing words with aligned (A; STAD = 0) vs. misaligned (M; STAD > 0.25) tokens and syllables.
        For each model and layer, the best $R^2$ is presented in boldface.
        A one-sided $t$-test is performed between the A and M splits at each layer to evaluate whether aligned words yield significantly better performance, with significance levels denoted as * ($p < 0.05$), ** ($p < 0.01$), and *** ($p < 0.001$). Results for more models can be found in the Appendix.
    }
    \label{table:stad-more}
\end{table*}
\section{$R^2$ on G2P and syllable counting for More Models}
\label{app:more-models}
We report the $R^2$ of G2P and syllable counting probing for more models in \Cref{table:stad-more}.
All additional models exhibit similar trends to the main models discussed in the paper.

\section{LLM Usage}
LLMs were used to assist with text refinement and data formatting, but not for generating core
research content.

\section{License and Terms for Use of Artifacts}
\label{app:license}
We rely on publicly released datasets and pretrained models, used strictly under their original terms. Specifically, we use the \emph{CMU Pronouncing Dictionary} (BSD-2-Clause), the \emph{Google-10000 English} word list (MIT), \emph{MMLU} (MIT), and \emph{GSM8K} (MIT). We also use \emph{CogNet}, which is released under CC BY-NC-SA 4.0; accordingly, we treat its content as non-commercial and attribution-preserving and make no redistribution beyond what the license permits. For the \emph{OpenHermes 2.5} compilation, no single uniform license is specified; the maintainers note heterogeneous upstream licenses and a placeholder policy, so we handle it as research-only, cite sources, and do not redistribute any of its contents. For pretrained models, our usage adheres to each model’s license: \emph{BERT} (Apache-2.0), \emph{GPT-2} (MIT/Modified MIT), \emph{ByT5} (Apache-2.0), \emph{Mistral-7B} (Apache-2.0), \emph{Llama 3/3.1} (Meta’s Llama Community License), \emph{Falcon-7B} (Apache-2.0), \emph{BLOOM} (BigScience OpenRAIL-M with use-based restrictions), and \emph{Yi-1.5} (Apache-2.0). We do not attempt to relicense any third-party artifacts; we redistribute nothing beyond what is explicitly permitted; and any derivatives (e.g., analysis outputs or small synthetic examples) are shared only in ways compatible with the upstream terms. We encourage downstream users to review the original license pages before reusing any asset and to note that “responsible-AI” licenses (e.g., OpenRAIL-M) and community licenses (e.g., Llama) include behavioral or compatibility restrictions that may differ from permissive open-source terms.

\end{document}